%% file: main.tex
\documentclass[letterpaper,journal]{IEEEtran}
\usepackage{amsmath,amsfonts,amssymb}
\usepackage{algorithmic}
\usepackage{algorithm}
\usepackage{array}
\usepackage[caption=false,font=normalsize]{subfig}
\usepackage{textcomp}
\usepackage{stfloats}
\usepackage{placeins}
\usepackage{url}
\usepackage{verbatim}
\usepackage{graphicx}
\usepackage{cite}
\usepackage{caption}
\usepackage{cleveref}
\crefname{figure}{Fig.}{Figs.}
\Crefname{figure}{Fig.}{Figs.}
\crefname{table}{Table}{Tables}
\Crefname{table}{Table}{Tables}
\usepackage{booktabs}
\usepackage{makecell}
\begin{document}

\title{\fontsize{18}{28}\selectfont PolarVis: A Dataset and Diffusion Framework for RGB-to-Polarization Perception in Adverse Driving Scenes}

\author{Ruilin Huang, Jue Wang, Zhen Tian, \IEEEmembership{Member, IEEE}, Xia Hua, F.~Richard Yu, \IEEEmembership{Fellow, IEEE}
\thanks{This work was supported in part by the Young Scientists Fund of the National Natural Science Foundation of China under Grant 62301305, in part by the General Program of the National Natural Science Foundation of China under Grant 62671366, and in part by the Guangdong Basic and Applied Basic Research Foundation under Grant 2025A1515012083. (Corresponding author: Xia Hua.)}
\thanks{Ruilin Huang is with the School of Communication and Information Engineering, Shanghai University, Shanghai 200444, China (e-mail: huang3132215071@163.com).}
\thanks{Jue Wang is with the School of Communication and Information Engineering, Shanghai University, Shanghai 200444, China (e-mail: wangjue0801@shu.edu.cn).}
\thanks{Zhen Tian is with the Guangdong Laboratory of Artificial Intelligence and Digital Economy (SZ), Shenzhen 518107, China (e-mail: tianzhen@gml.ac.cn).}
\thanks{F. Richard Yu is with the Department of Electrical Engineering, Carleton University, Ottawa, Ontario K1S 5B6, Canada (e-mail: richardyu@carleton.ca).}
\thanks{Xia Hua is with the School of Communication and Information Engineering, Shanghai University, Shanghai 200444, China (e-mail: xiahua@shu.edu.cn).}}


\maketitle

\begin{figure*}[!t]
	\centering
	\begin{minipage}[t]{\textwidth}
		\centering
		\includegraphics[width=\linewidth]{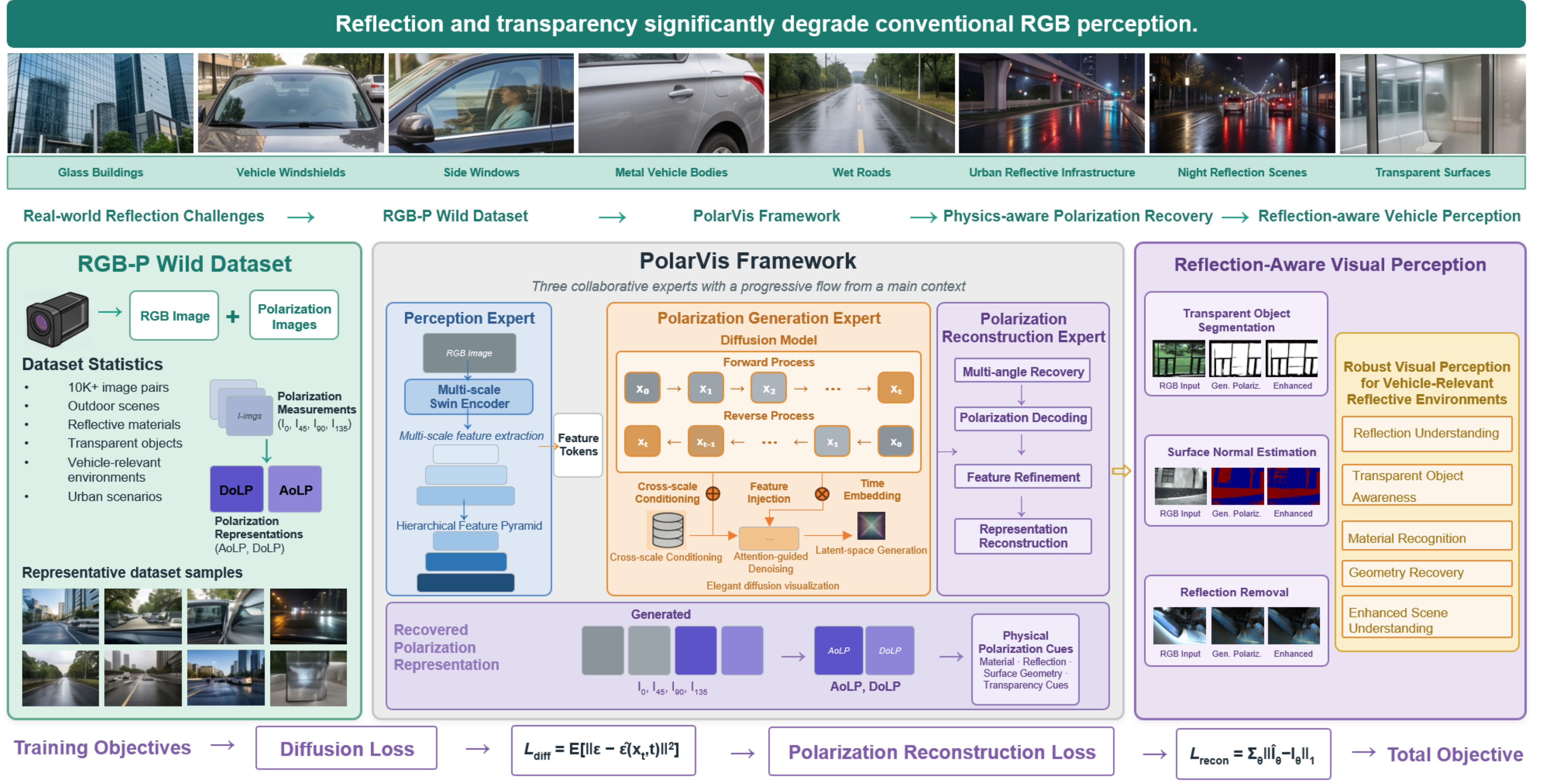}
		\captionof{figure}{Overview of PolarVis. Reflective and transparent surfaces commonly encountered in vehicle-relevant environments can significantly impair conventional RGB-based visual perception. We first construct RGB-P Wild, a large-scale RGB-polarization dataset containing diverse outdoor reflective scenes. Leveraging this dataset, PolarVis employs a diffusion-based RGB-to-polarization generation framework to recover polarization representations, including multi-angle polarization images and their corresponding AoLP and DoLP maps. These recovered polarization cues provide complementary information about material properties, reflections, surface geometry, and transparency, which can facilitate reflection-aware visual perception tasks such as reflection removal, transparent object segmentation, surface normal estimation, and 3D reconstruction.}
		\label{fig1}
	\end{minipage}
\end{figure*}

\begin{abstract}
	Reflections, glare, transparent structures, and material-dependent appearance variations can introduce ambiguity into RGB-based visual perception in vehicle-relevant environments. Polarization provides complementary reflection- and material-sensitive information, but its acquisition requires dedicated sensing hardware. This paper presents RGB-P Wild, a real-world dataset containing 10,000 paired RGB and four-angle linear-polarization image sets, together with PolarVis, a diffusion-based framework for recovering polarization representations from a single RGB image. PolarVis combines hierarchical RGB feature extraction, diffusion-based feature refinement, four angle-specific decoders, and multi-scale attention recalibration to reconstruct polarization intensity images at $0^\circ$, $45^\circ$, $90^\circ$, and $135^\circ$. On RGB-P Wild, PolarVis achieves a normalized-domain MSE of 0.0465, a PSNR of 27.83~dB, and an SSIM of 0.8597. Compared with the released PolarAnything model under the adopted evaluation protocol, PolarVis obtains an AoLP error of $25.66^\circ$ and a DoLP MAE of 0.0404, compared with $46.03^\circ$ and 0.1084, respectively. Cross-dataset evaluation on vehicle-mounted imagery further examines transfer beyond the training domain. In downstream evaluation, a frozen PCDNet achieves AP50/AP50:95 of 86.3\%/59.1\% on 990 RGBP-Car test images using PolarVis-generated polarization, compared with 87.0\%/59.8\% using measured polarization. These results indicate that RGB-derived polarization retains useful polarimetric information for the evaluated perception tasks, while remaining distinct from calibrated physical polarization measurements.
\end{abstract}


\begin{figure*}[!t]
	\centering
	\begin{minipage}[t]{\textwidth}
		\centering
		\includegraphics[width=\linewidth]{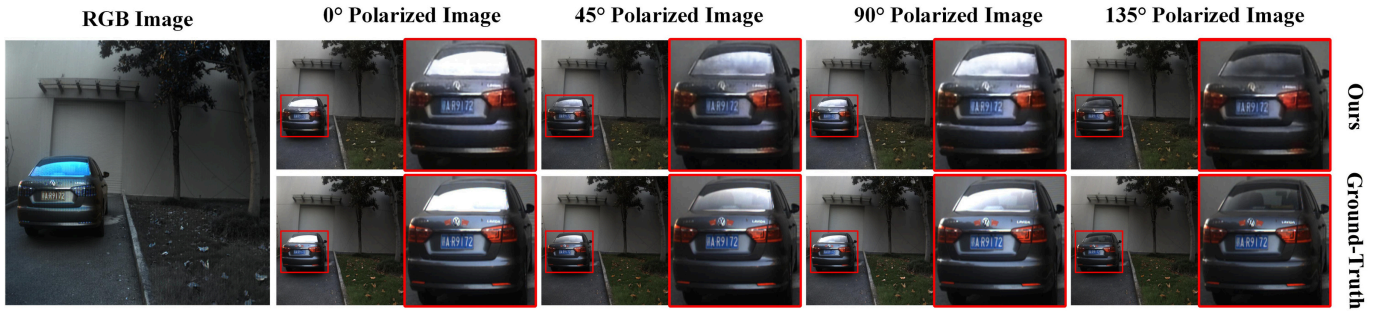}
		\captionof{figure}{Qualitative polarization generation results of PolarVis on RGB-P Wild. Given a single RGB image, PolarVis estimates four linear polarization images at $0^\circ$, $45^\circ$, $90^\circ$, and $135^\circ$. The generated polarization images are visually compared with the corresponding ground-truth polarization observations, showing that the proposed framework can reconstruct the main reflection- and material-related structures across different polarization angles.}
		\label{fig2}
	\end{minipage}
\end{figure*}

\begin{figure}[h]
	\centering
	\subfloat[]{%
		\includegraphics[width=0.48\linewidth]{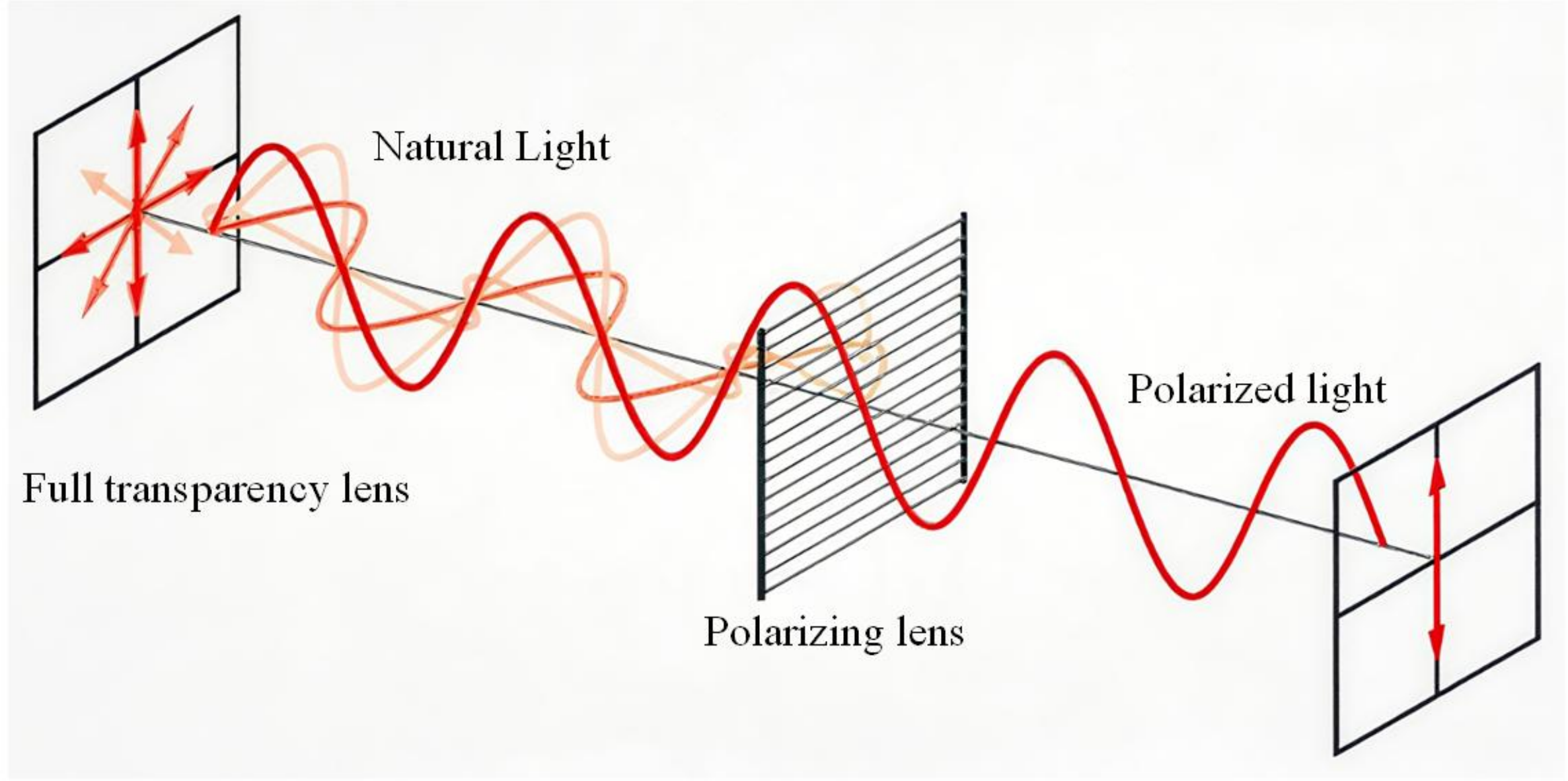}%
		\label{fig3a}%
	}%
	\hfill%
	\subfloat[]{%
		\includegraphics[width=0.24\linewidth]{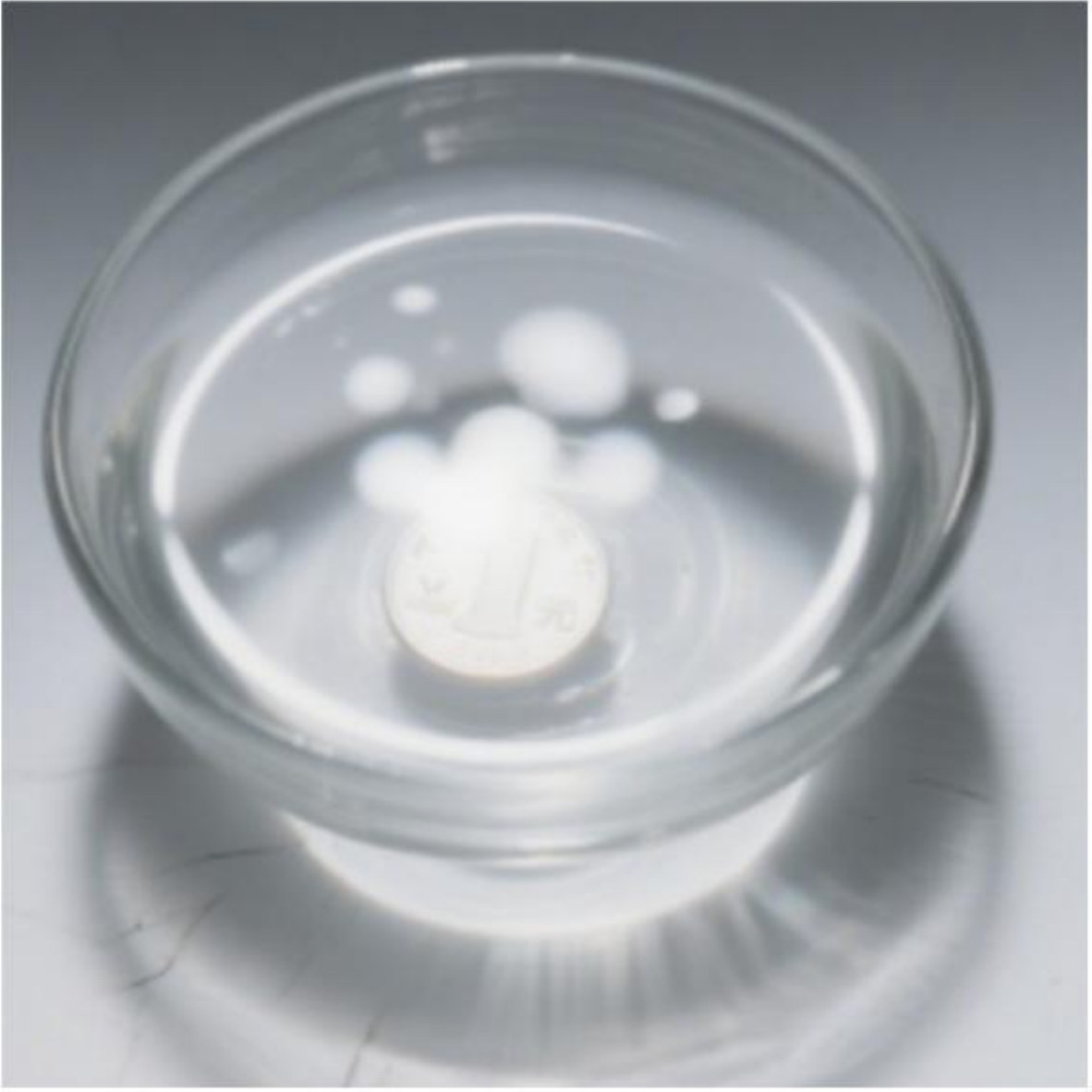}%
		\label{fig3b}%
	}%
	\hfill%
	\subfloat[]{%
		\includegraphics[width=0.24\linewidth]{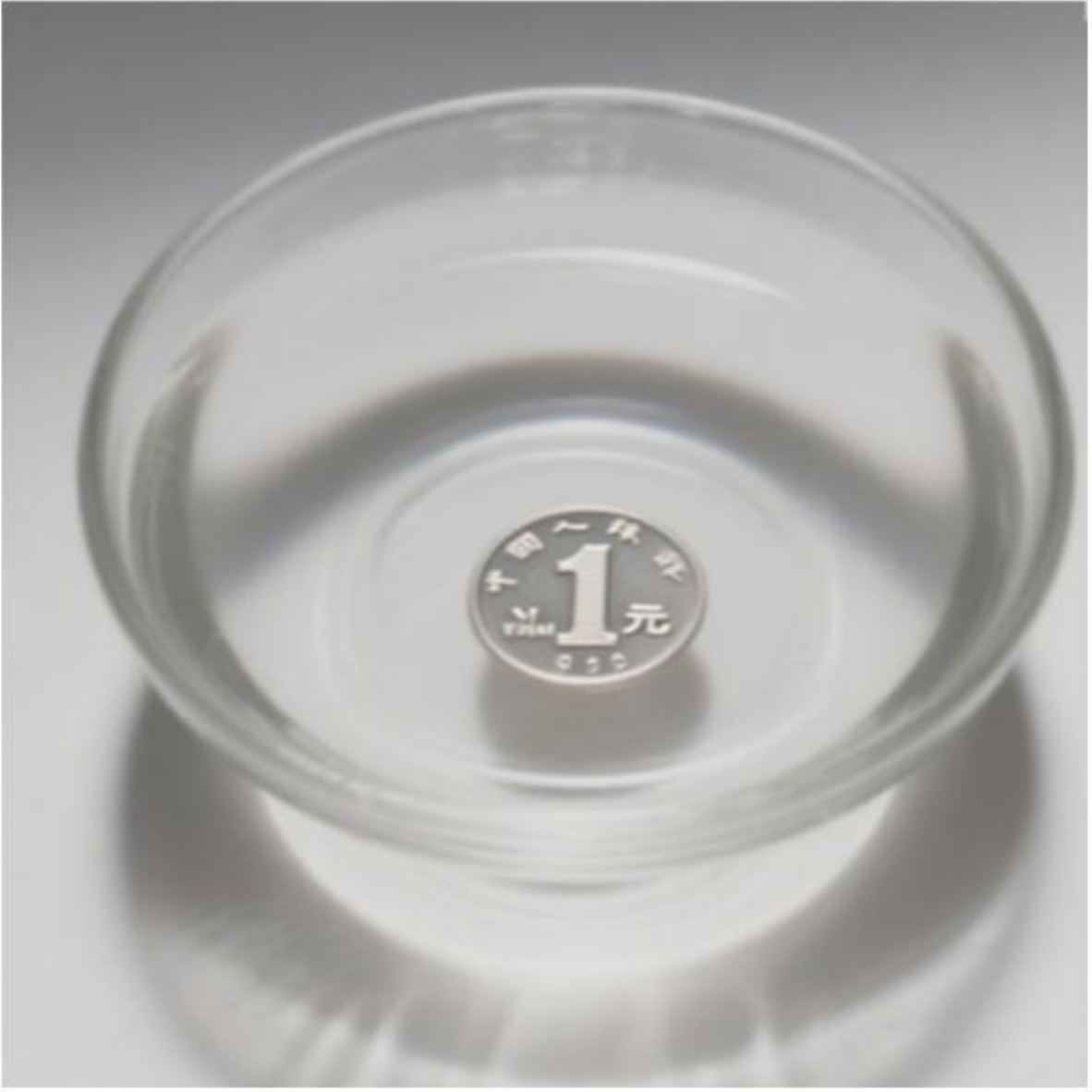}%
		\label{fig3c}%
	}%

	\vspace{0.5em}%
	
	\subfloat[]{%
		\includegraphics[width=0.24\linewidth]{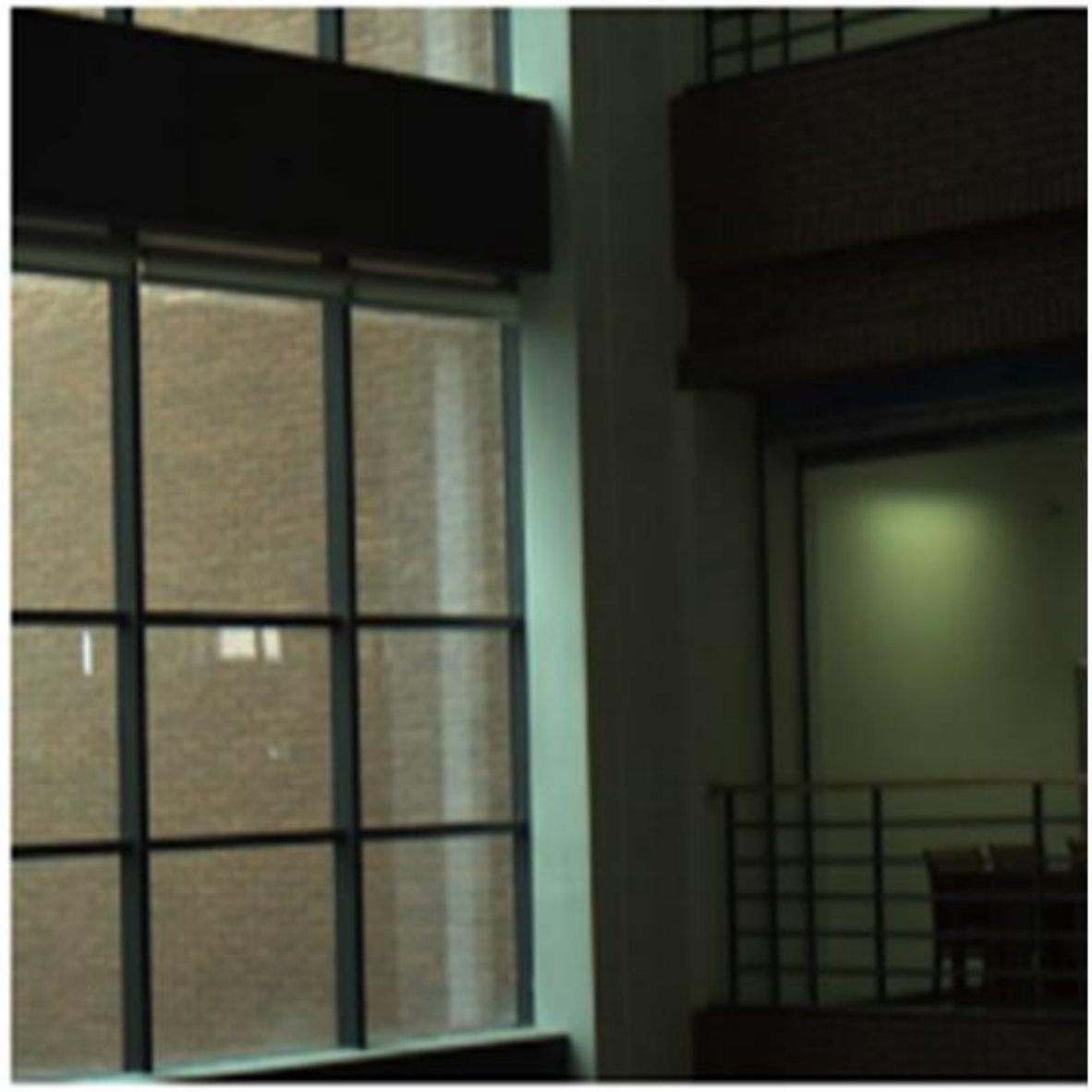}%
		\label{fig3d}%
	}%
	\hfill%
	\subfloat[]{%
		\includegraphics[width=0.24\linewidth]{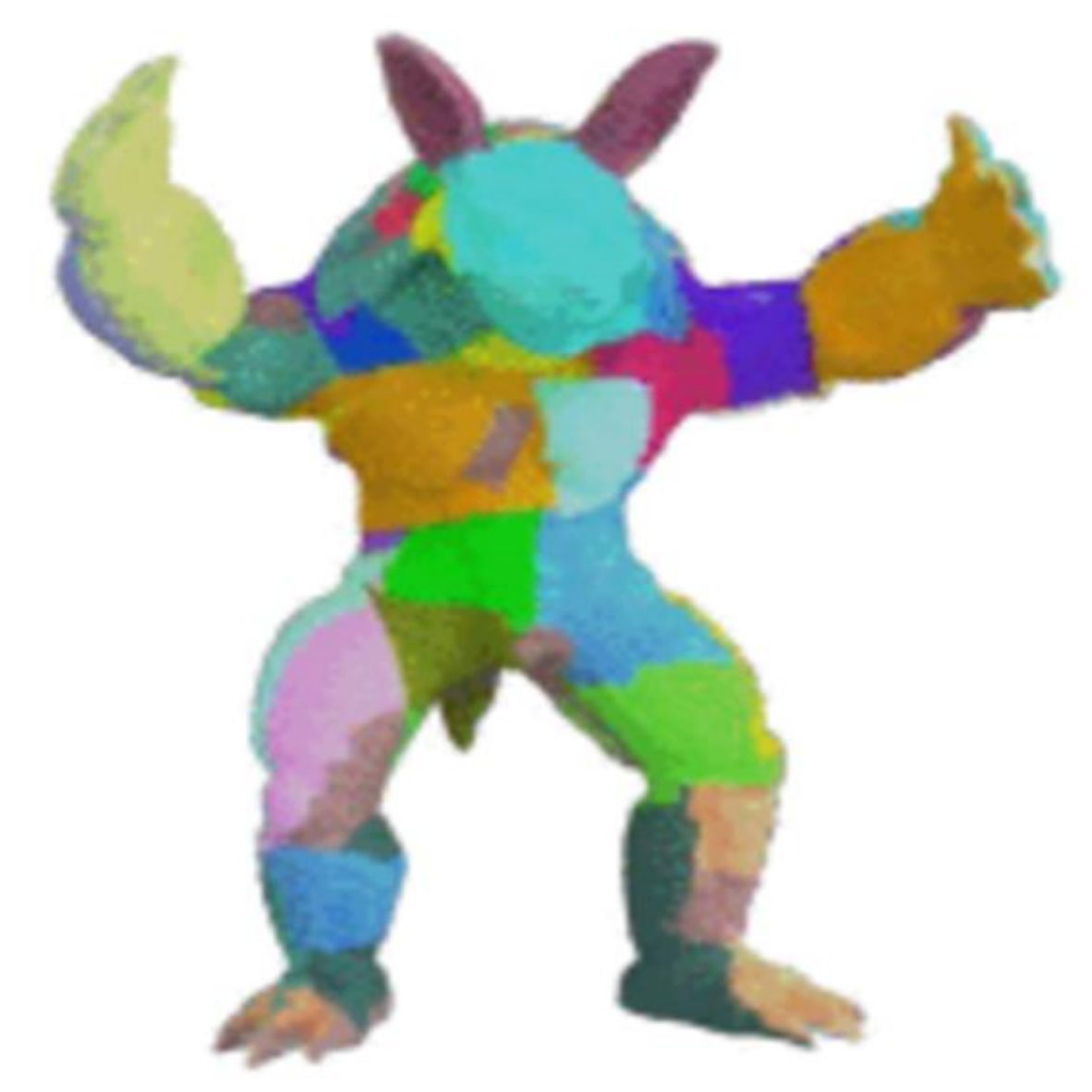}%
		\label{fig3e}%
	}%
	\hfill%
	\subfloat[]{%
		\includegraphics[width=0.24\linewidth]{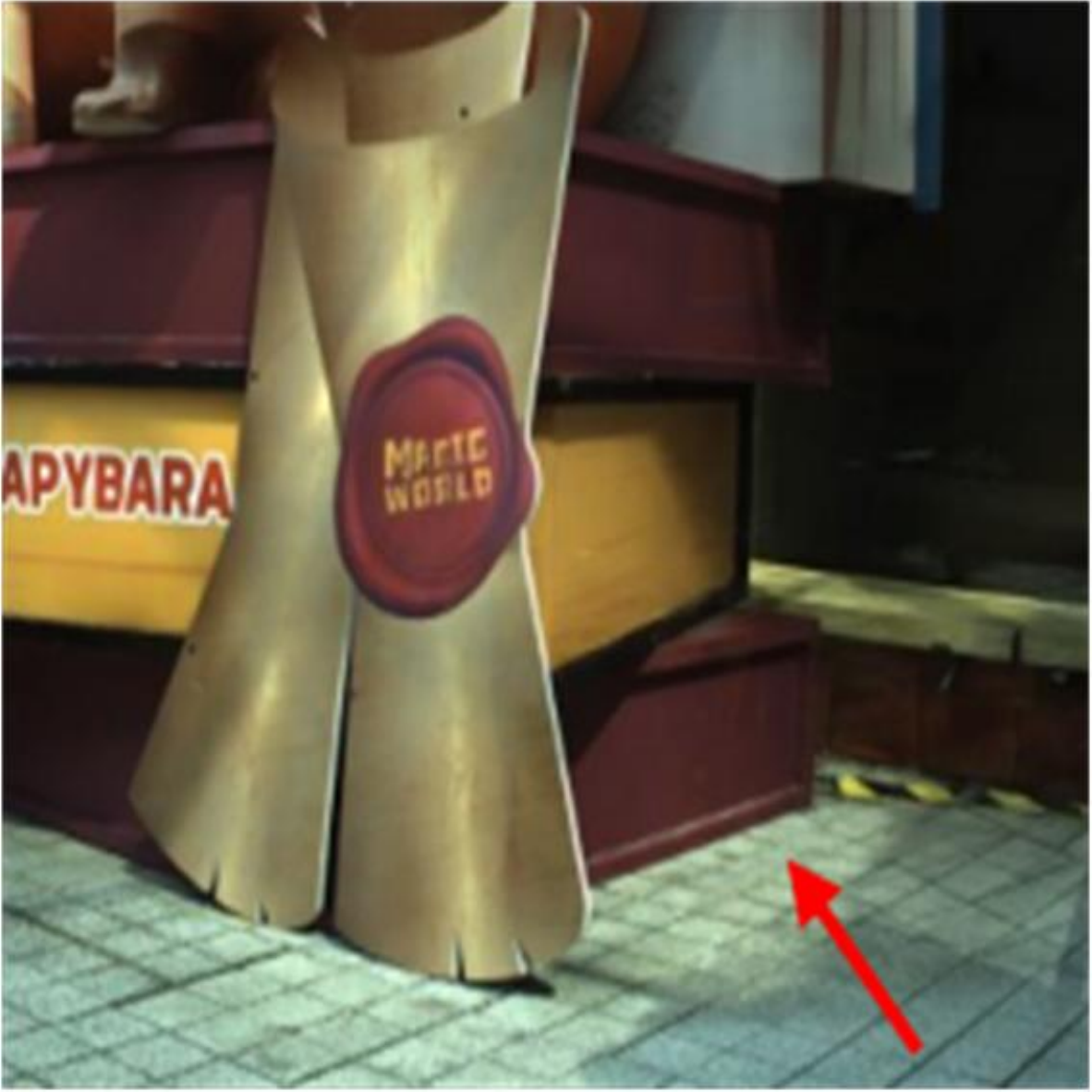}%
		\label{fig3f}%
	}%
	\hfill%
	\subfloat[]{%
		\includegraphics[width=0.24\linewidth]{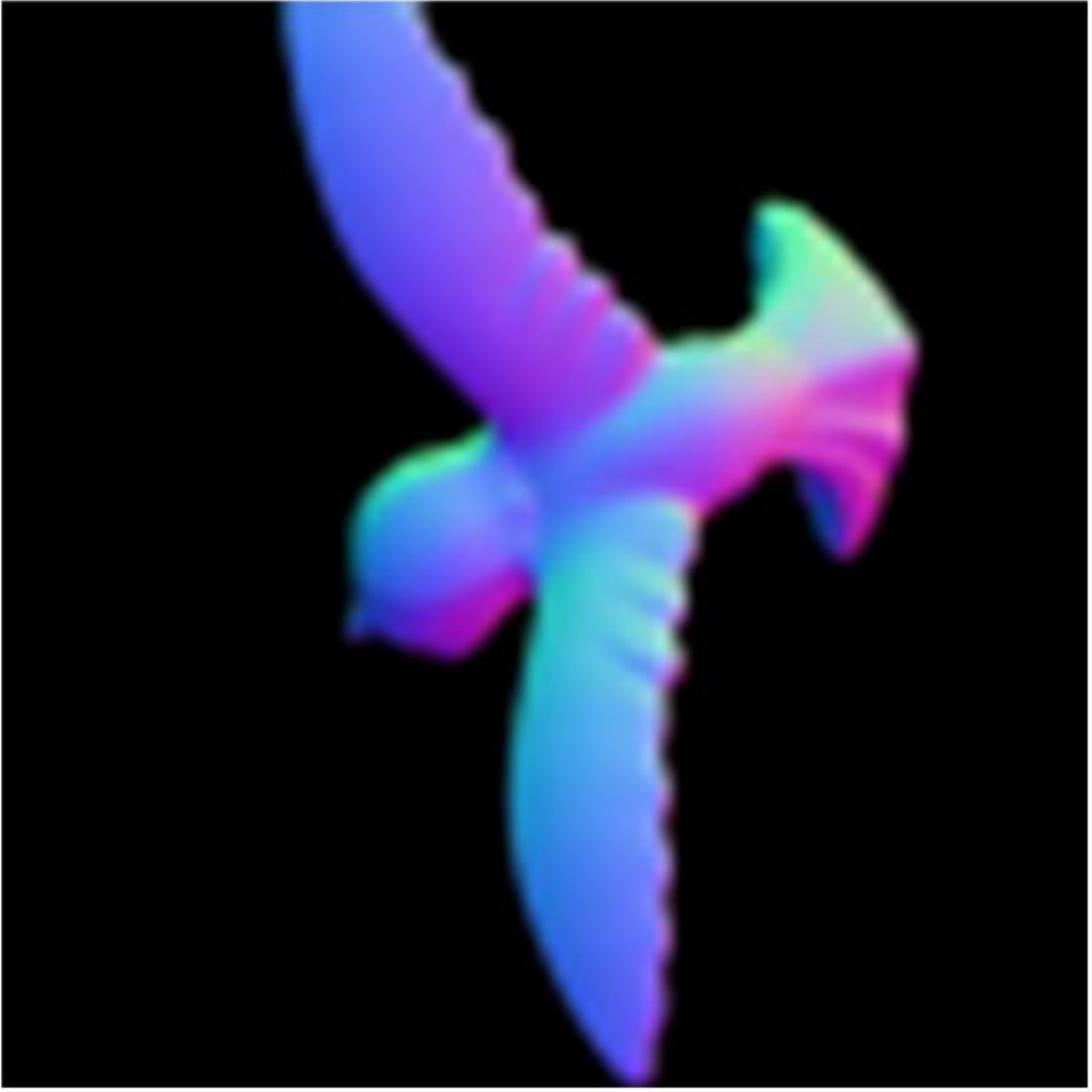}%
		\label{fig3g}%
	}
	\caption{Physical basis and representative applications of polarization imaging. (a) illustrates light--surface interaction; (b)--(c) compare RGB and polarization observations; and (d)--(g) summarize applications reported in prior work, including transparent-object segmentation with PGSNet \cite{Alpher06}, 3D reconstruction with Polarimetric MVIR \cite{Alpher51}, reflection removal with PolarFree \cite{Alpher47}, and surface-normal estimation with TransSfP \cite{Alpher11}. These examples provide application context and are not experimental results of PolarVis.}
	\label{fig3}
\end{figure}

\section{Introduction}

\IEEEPARstart{R}{obust} visual perception is essential for intelligent vehicles. Wet-road glare, windshields, metallic reflections, transparent structures, and low contrast can obscure boundaries and alter material appearance in RGB images \cite{Alpher71,Alpher72,Alpher75,Alpher76}. As illustrated in \Cref{fig1}, PolarVis predicts four-angle polarization images from an RGB input and derives the angle of linear polarization (AoLP) and degree of linear polarization (DoLP) for subsequent evaluation.

Polarization provides information beyond intensity and color because its state depends on surface reflection, material properties, scattering, and local geometry \cite{Alpher44}. As illustrated in \Cref{fig3}, polarization has supported reflection-, transparency-, and geometry-related perception tasks in prior studies \cite{Alpher06,Alpher11,Alpher47,Alpher52}.

Dedicated polarization sensors increase hardware cost, calibration effort, and processing complexity \cite{Alpher65,Alpher70}. RGB-to-polarization generation offers a software alternative, and diffusion models can learn this structured cross-modal mapping \cite{Alpher14,Alpher16,Alpher22,Alpher23}.

We construct RGB-P Wild with paired RGB and four-angle polarization images under diverse illumination and reflective conditions. Compared with existing datasets \cite{Alpher01,Alpher05,Alpher06,Alpher11,Alpher31,Alpher33}, it provides larger-scale coverage of materials and outdoor conditions relevant to vehicle perception, although it is not vehicle-mounted. We therefore evaluate transfer on the GM dataset \cite{Alpher43} and vehicle detection on RGBP-Car \cite{Alpher83}. Representative generated images are shown in \cref{fig2}.

The contributions are threefold. First, RGB-P Wild contains 10,000 paired RGB and four-angle polarization image sets. Second, PolarVis combines hierarchical encoding, diffusion-based refinement, angle-specific decoding, and multi-scale attention recalibration. Third, we evaluate reconstruction quality, polarimetric fidelity, computational efficiency, cross-dataset transfer, transparent-object segmentation, and frozen vehicle detection to examine whether the generated polarization cues retain useful information for downstream perception.

\section{Related Work}

\subsection{Vehicle-Oriented Visual Perception Under Challenging Conditions}

Vehicle perception must operate under adverse weather, low illumination, reflective surfaces, wet roads, and sensor degradation \cite{Alpher71,Alpher72,Alpher75,Alpher76}. Recent work has also investigated environmental reasoning for adaptive target tracking under changing sensing conditions \cite{tian2026cater}. Reflection-related ambiguity affects three-dimensional sensing and mapping \cite{Alpher50}. Complementary sensing and enhancement methods use RGB-thermal fusion, infrared, event, range-gated, and polarization modalities \cite{Alpher30,Alpher72,Alpher74,Alpher77,Alpher78}. RGB-polarimetric sensing has been investigated for driving perception \cite{Alpher43} and robust car detection \cite{Alpher83}. Unlike methods that add a physical sensor stream, PolarVis generates polarization-aware cues from RGB inputs for use with existing visual pipelines \cite{Alpher76,Alpher79,Alpher80}.

\subsection{Polarization Imaging for Reflection- and Geometry-Aware Perception}

Polarization is related to reflection, scattering, material properties, and geometry \cite{Alpher44}. It has been applied to event-assisted polarization reconstruction \cite{Alpher04}, transparent-object segmentation \cite{Alpher06,Alpher09,Alpher10,Alpher57}, shape-from-polarization and normal estimation \cite{Alpher03,Alpher05,Alpher08,Alpher11,Alpher37,Alpher45,Alpher46}, depth enhancement \cite{Alpher02,Alpher12}, color constancy \cite{Alpher07}, semantic segmentation \cite{Alpher53}, reflection removal and separation \cite{Alpher34,Alpher47,Alpher48,Alpher49}, and descattering and restoration \cite{Alpher17,Alpher52,Alpher54,Alpher55,Alpher60}. Passive polarized vision is also relevant to mobile robots and autonomous vehicles \cite{Alpher73}.

\subsection{RGB-Polarization Datasets and Polarization Synthesis}

Existing datasets include Spectro-Polarimetric Dataset \cite{Alpher01,Alpher32,Alpher34}, SPW \cite{Alpher05}, MCubeS \cite{Alpher31}, RGB-P Glass \cite{Alpher06}, PKU-PI \cite{Alpher33}, and synthetic object-level data \cite{Alpher42}. Sparse polarization sensors and focal-plane demosaicking further support aligned acquisition \cite{Alpher69,Alpher70}. For RGB-to-polarization synthesis, Lin et al. established an estimation benchmark \cite{Alpher68}, while PolarAnything introduced diffusion-based polarimetric synthesis \cite{Alpher65}. RGB-P Wild provides larger-scale paired real-world samples for learning multi-angle generation under diverse outdoor conditions \cite{Alpher77}.

\subsection{Diffusion-Based Cross-Modal Generation}

Diffusion models support image synthesis and conditional generation \cite{Alpher14,Alpher18,Alpher19,Alpher20,Alpher22,Alpher23,Alpher25,Alpher60}, with deterministic sampling and quantization improving efficiency \cite{Alpher13,Alpher15}. Beyond conventional RGB synthesis, diffusion priors have also been adapted to hyperspectral image restoration and super-resolution \cite{jia2024diffhsr}. PolarVis conditions latent diffusion \cite{Alpher16} on Swin Transformer features \cite{Alpher35} to predict four angle-specific images. PolarAnything similarly applies pretrained diffusion priors to polarimetric image synthesis \cite{Alpher65}.

\begin{figure*}[h]
	\centering
	\begin{minipage}[t]{\textwidth}
		\centering
		\includegraphics[width=\linewidth]{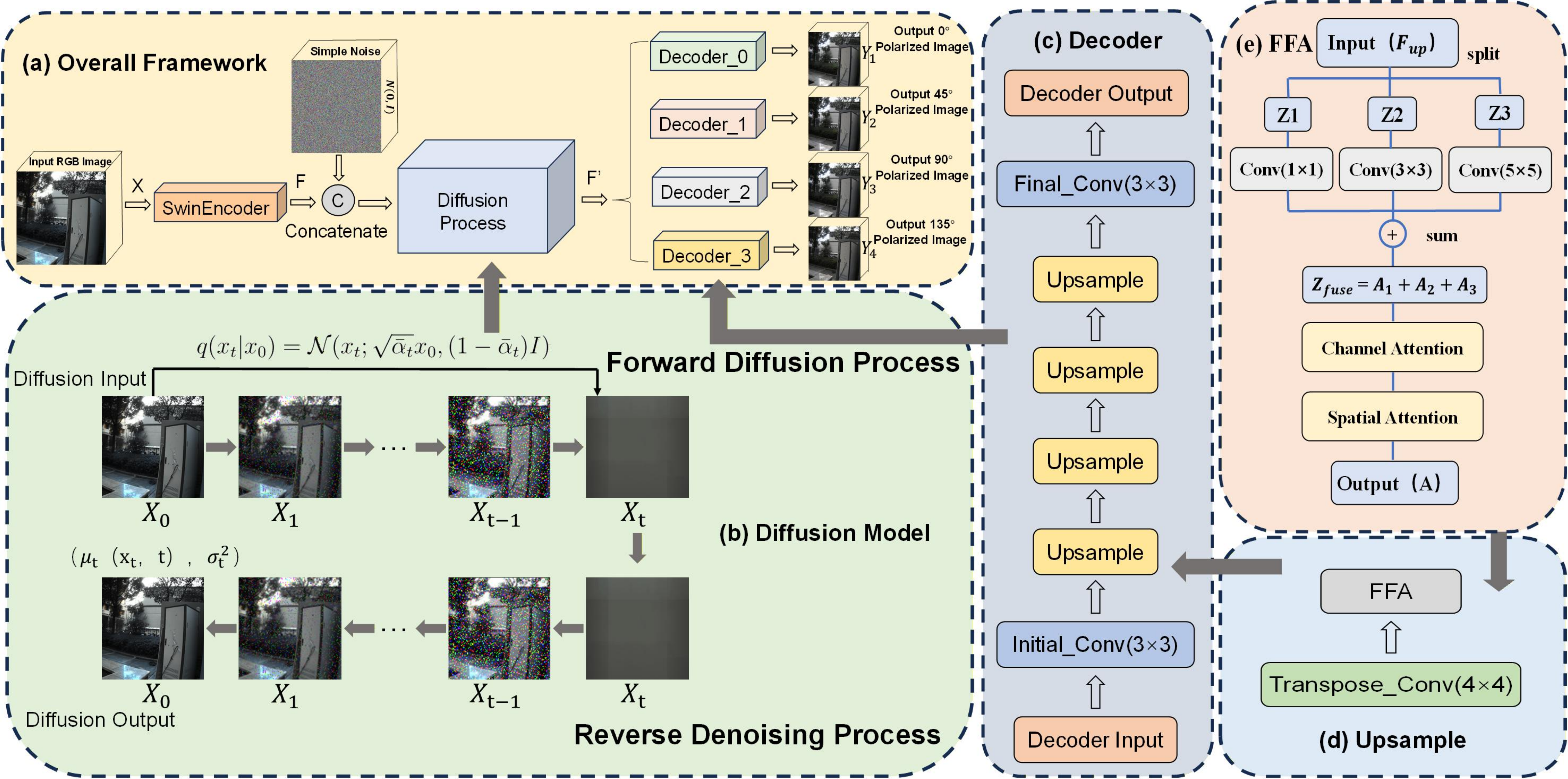}
		\caption{Detailed architecture of PolarVis. (a) Overall RGB-to-polarization generation pipeline, including the Swin-Transformer-based feature encoder, optimized diffusion module, and four angle-specific polarization decoders. (b) Illustration of the forward diffusion and reverse denoising processes. (c) Decoder data flow for polarization image reconstruction. (d) Upsampling module in the decoder. (e) Multi-Scale Feature Fusion and Attention Feature Recalibration (FFA) module used to enhance reconstructed polarization features.}
		\label{fig4}
	\end{minipage}
\end{figure*}

\section{Method}

\subsection{Overview}

This study proposes a diffusion-based framework that synthesizes multi-angle linear polarization images from a single RGB input. From the perspective of vehicle-oriented perception, the RGB input can be regarded as a standard image frame captured by a conventional camera, while the generated polarization images provide auxiliary reflection- and material-sensitive cues for subsequent perception models \cite{Alpher79,Alpher80}. As depicted in \cref{fig4}, the pipeline contains three core components: a Swin-Transformer-based encoder, an optimized diffusion module, and multiple enhanced decoders.

For an input RGB image~$X$, the framework first employs a Swin-Transformer encoder to extract hierarchical features~$F$:

\begin{equation}
	F = \text{SwinEncoder}(X)
	\label{eq1.1}
\end{equation}

These features are then processed through a customized diffusion model:

\begin{equation}
	F' = \text{DiffusionModel}(F)
	\label{eq1.2}
\end{equation}

Finally, four specialized decoders generate linear polarization images at the canonical angles $0^\circ$, $45^\circ$, $90^\circ$, and $135^\circ$:

\begin{equation}
	Y_i = \text{Decoder}_i(F'), i \in \{1,2,3,4\}.
	\label{eq1.3}
\end{equation}

\subsection{Swin-Transformer-based Encoder}

Taking inspiration from the excellent feature extraction capabilities demonstrated in Swin Transformer\cite{Alpher35}, we developed an encoder based on its architecture. The encoder gradually extracts multi-scale features through a series of key components to ensure effective representation of input data at various resolutions.

\noindent\textbf{Patch Embedding.} The initial stage transforms the input image into non-overlapping patches. For patches of size $P \times P$, we first flatten each $P \times P \times 3$ patch into a vector of dimension $3P^2$, then map it to a D-dimensional embedding space via a linear projection:

\begin{equation}
	\text{PatchEmbed}(X) = \text{Conv2d}(X, P, P) \in \mathbb{R}^{\frac{H}{P} \times \frac{W}{P} \times D}
	\label{eq1.4}
\end{equation}

\noindent where $X \in \mathbb{R} ^ {H \times W \times C}$ denotes the input image, P represents the patch size, and D is the embedding dimension.

\noindent\textbf{Window-based Self-Attention.} We adopt the efficient window-based self-attention mechanism from Swin Transformer, which computes attention within local windows of size $M \times M$ (M=7 in our implementation). The multi-head self-attention with relative position bias $B$ follows Swin Transformer \cite{Alpher35} and is computed as:

\begin{equation}
	\text{Attention}(Q, K, V) = \text{SoftMax}(\frac{QK^T}{\sqrt{d}} + B)V
	\label{eq1.5}
\end{equation}

\noindent where Q, K, and V represent query, key, and value matrices respectively, d is the key dimension, and B encodes relative positional information.

\noindent\textbf{Progressive Downsampling.} The encoder employs a hierarchical structure with progressive feature downsampling. At each stage boundary, four spatially adjacent 2 $\times$ 2 patches are concatenated along the channel dimension, resulting in a new patch with quadrupled channel depth but halved spatial resolution.This merging strategy preserves information density while enabling multi-scale representation learning.

\subsection{Optimized Diffusion Model}

\indent Our diffusion model implements a Markov chain process with T time steps, carefully balancing computational efficiency with generation quality. The noise schedule $\beta_1, \ldots, \beta_t$\cite{Alpher36} is linearly increased from $10^{-4}$ to $0.02$ over 100 time steps, with signal preservation ratio $\alpha_t = 1 - \beta_t$. The forward process follows a Gaussian transition:

\begin{equation}
	q(x_t|x_0) = \mathcal{N}(x_t; \sqrt{\bar{\alpha}_t}x_0, (1-\bar{\alpha}_t)I)
	\label{eq1.6}
\end{equation}

\noindent where $\bar{\alpha}_t = \prod_{s=1}^t \alpha_s$. The noisy sample $x_t$ is obtained through:

\begin{equation}
	x_t = \sqrt{\bar{\alpha}_t}x_0 + \sqrt{1-\bar{\alpha}_t}\epsilon, \epsilon \sim \mathcal{N}(0, I)
	\label{eq1.7}
\end{equation}

\noindent The reverse process estimates posterior mean and variance:

\begin{equation}
	\mu_t(x_t, t) = \frac{1}{\sqrt{\alpha_t}}(x_t - \frac{\beta_t}{\sqrt{1-\bar{\alpha}_t}}\epsilon_\theta(x_t, t))
	\label{eq1.8}
\end{equation}

\begin{equation}
	\sigma^2_t = \beta_t\frac{1-\bar{\alpha}_{t-1}}{1-\bar{\alpha}_t}
	\label{eq1.9}
\end{equation}

\noindent We incorporate temporal embeddings to enhance the model's understanding of different diffusion stages.

\subsection{Enhanced Decoder Architecture}

\indent Our framework deploys four structurally identical decoders, each dedicated to generating polarization images corresponding to one of the four canonical angles ($0^\circ$, $45^\circ$, $90^\circ$, $135^\circ$). The decoder architecture is designed to reconstruct high-fidelity polarization states by integrating multi-scale contextual cues and adaptive feature recalibration. As illustrated in \cref{fig4}, each decoder comprises a progressive upsampling stage followed by a unified Multi-Scale Feature Fusion and Attention Feature Recalibration Module (FFA).

\noindent\textbf{Progressive Upsampling.} The upsampling stage incrementally increases the spatial resolution of the input latent feature $F'$ via transposed convolutions. Specifically, each upsampling block first applies a $3 \times 3$ convolution, followed by batch normalization, ReLU activation, and a transposed convolution operation:

\begin{equation}
	F_{\text{up}} = \text{ReLU}\left(\text{BN}\left(\text{ConvTranspose2d}\left(\text{Conv}_{3\times3}(F')\right)\right)\right)
	\label{eq:1.15}
\end{equation}

\noindent The resulting feature $F_{\text{up}}$ serves as the input to the subsequent FFA module.

\noindent\textbf{Multi-scale feature fusion and attention feature recalibration module (FFA).} The FFA module is designed to enhance the quality and richness of feature reconstruction. To boost representational capacity, the FFA module first splits the input feature $F_{\text{up}} \in \mathbb{R}^{H \times W \times C}$ equally along the channel dimension into three sub-features:

\begin{equation}
	A_1, A_2, A_3 = \text{Split}(F_{\text{up}}, \text{num\_splits}=3, \text{dim}=1)
	\label{eq:1.14}
\end{equation}

Each sub-feature is then processed by a dedicated convolutional branch with distinct receptive fields to capture multi-scale patterns:

\begin{itemize}
	\item $A_1$ is processed by a $1 \times 1$ convolution to enable efficient cross-channel information interaction;
	\item $A_2$ is convolved with a $3 \times 3$ kernel to model local spatial contextual information;
	\item $A_3$ undergoes a $5 \times 5$ convolution to capture global structural context.
\end{itemize}

The outputs of these branches are denoted as $Z_1, Z_2, Z_3$, respectively, and aggregated via element-wise summation to form a fused multi-scale feature:

\begin{equation}
	Z_{\text{fuse}} = Z_1 + Z_2 + Z_3
	\label{eq1.16}
\end{equation}

Subsequently, a hybrid attention mechanism—combining channel and spatial attention—is applied to $Z_{\text{fuse}}$ for feature response recalibration. The channel attention weight $w_c$ is computed by aggregating features through average and max pooling, followed by a shared multi-layer perceptron (MLP):

\begin{equation}
	w_c = \sigma\left(f_c\left(\text{AvgPool}(Z_{\text{fuse}})\right) + f_c\left(\text{MaxPool}(Z_{\text{fuse}})\right)\right)
	\label{eq1.11}
\end{equation}

The spatial attention weight $w_s$ is generated by concatenating average and max pooled features along the channel axis, followed by a $7 \times 7$ convolution:

\begin{equation}
	w_s = \sigma\left(\text{Conv}{7 \times 7}\left(\left[\text{AvgPool}(Z{\text{fuse}});\ \text{MaxPool}(Z_{\text{fuse}})\right]\right)\right)
	\label{eq1.12}
\end{equation}

The final attention-modulated feature $A$ is obtained by sequentially applying channel and spatial attention weights to $Z_{\text{fuse}}$:

\begin{equation}
	A = \left(Z_{\text{fuse}} \otimes w_c\right) \otimes w_s
	\label{eq1.13}
\end{equation}

where $\otimes$ denotes channel-wise or spatial broadcasting multiplication, and $\sigma$ represents the sigmoid activation function.

This refined feature $A$ is either passed to the next upsampling stage or directly used to generate the polarization image output $Y_i$ for angle $i$. The resulting feature is used for angle-specific polarization reconstruction.

\subsection{Loss Functions}

\indent Our training objective combines diffusion and reconstruction losses:

\noindent\textbf{Diffusion Loss.} The primary loss function for the diffusion model is defined as follows:

\begin{equation}
	\mathcal{L}_{\text{diff}} = \mathbb{E}_{t,\epsilon}[\|\epsilon - \epsilon_\theta(x_t,t)\|^2]
	\label{eq1.17}
\end{equation}

\noindent This loss measures the discrepancy between the model-predicted noise $\epsilon_\theta(x_t, t)$ and the true noise $\epsilon$. By minimizing this loss, the model learns to accurately reconstruct clean images at different time steps t, effectively guiding the denoising process.

\noindent\textbf{Reconstruction Loss.} The function definition of the reconstruction loss (smoothed L1 loss) is as follows:

\begin{equation}
	\mathcal{L}_{\text{recon}} = \text{SmoothL1Loss}(Y_{\text{pred}}, Y_{\text{true}})
	\label{eq1.18}
\end{equation}

\noindent This loss function is used to evaluate the difference between the generated image $Y_{\text{pred}}$ and the real image $Y_{\text{true}}$. Compared with the standard L1 loss, Smooth L1 loss is less sensitive to outliers and helps stabilize training while preserving local structural details in the reconstructed polarization images.

\noindent\textbf{Total Loss.} In the overall loss function, the hyperparameter $\lambda$ is used to adjust the weight of the diffusion model loss and reconstruction loss to balance their impact on model training:

\begin{equation}
	\mathcal{L}_{\text{total}} = \mathcal{L}_{\text{diff}} + \lambda\mathcal{L}_{\text{recon}}
	\label{eq1.19}
\end{equation}

\noindent where $\lambda$ balances the contribution of each loss component. The design of this combined loss function not only focuses the model on effectively removing noise but also ensures that the generated image visually approximates the real image, thereby enhancing the robustness of the model and the overall quality of the generated image.

\subsection{Training and Inference}

PolarVis is trained on RGB-P Wild. At inference, RGB features are perturbed and refined using $K=10$ denoising updates by default, before four independently parameterized decoders produce the $0^\circ$, $45^\circ$, $90^\circ$, and $135^\circ$ images. The framework follows feature-space conditional generation \cite{Alpher16}; measured polarization is not required as input. The efficiency experiment additionally evaluates $K=1$ and $K=5$. The reduced-step procedure is an implementation-specific stochastic refinement, not standard DDIM sampling.

\section{Experiments and Analysis}

\subsection{RGB-P Wild Dataset}

\begin{figure}[h]
	\centering
	\includegraphics[width=\linewidth]{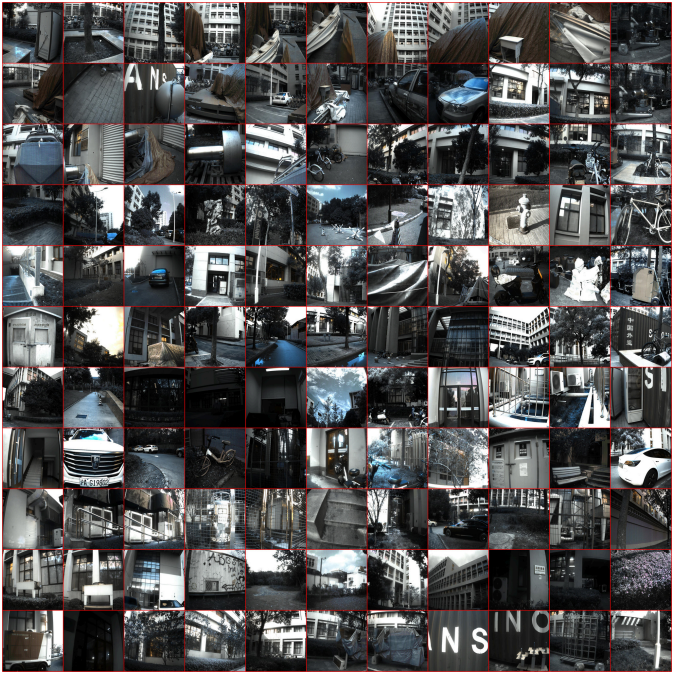}
	\caption{Sample overview of RGB-P Wild. 121 RGB images arranged in an $11 \times 11$ grid are selected from 10,000 images in total. The subset covers diverse illumination conditions (daytime/nighttime), object materials (glass, plastic, metal, liquid, etc.), and scene types (indoor/outdoor), reflecting the dataset's variability in environmental context, material property, and lighting condition.}
	\label{fig5}
\end{figure}

RGB-P Wild contains 10,000 image sets, each comprising one RGB image and four associated linear-polarization images at $0^\circ$, $45^\circ$, $90^\circ$, and $135^\circ$. As summarized in \Cref{tab1}, existing polarization datasets \cite{Alpher01,Alpher05,Alpher06,Alpher37} provide complementary scene and object coverage, whereas RGB-P Wild emphasizes diverse reflective materials and illumination conditions. The data are divided into 7,500 training sets and 2,500 evaluation sets. Representative RGB samples are shown in \cref{fig5}.

RGB-P Wild was not collected from a moving vehicle and does not contain vehicle-detection or road-semantic annotations. Its relevance to intelligent vehicles lies in optical challenges such as glass, metal, liquid regions, glare, and reflection-dominated appearance. We therefore use a separate vehicle dataset for direct downstream evaluation rather than treating RGB-P Wild as a driving benchmark.

\begin{figure*}[!t]
	\centering
	\subfloat[Color-polarization sampling of the LUCID TRI050S1-QC camera \cite{Alpher84}.\label{fig6a}]{%
		\begin{minipage}[b]{0.43\textwidth}
			\centering
			\includegraphics[width=0.75\linewidth]{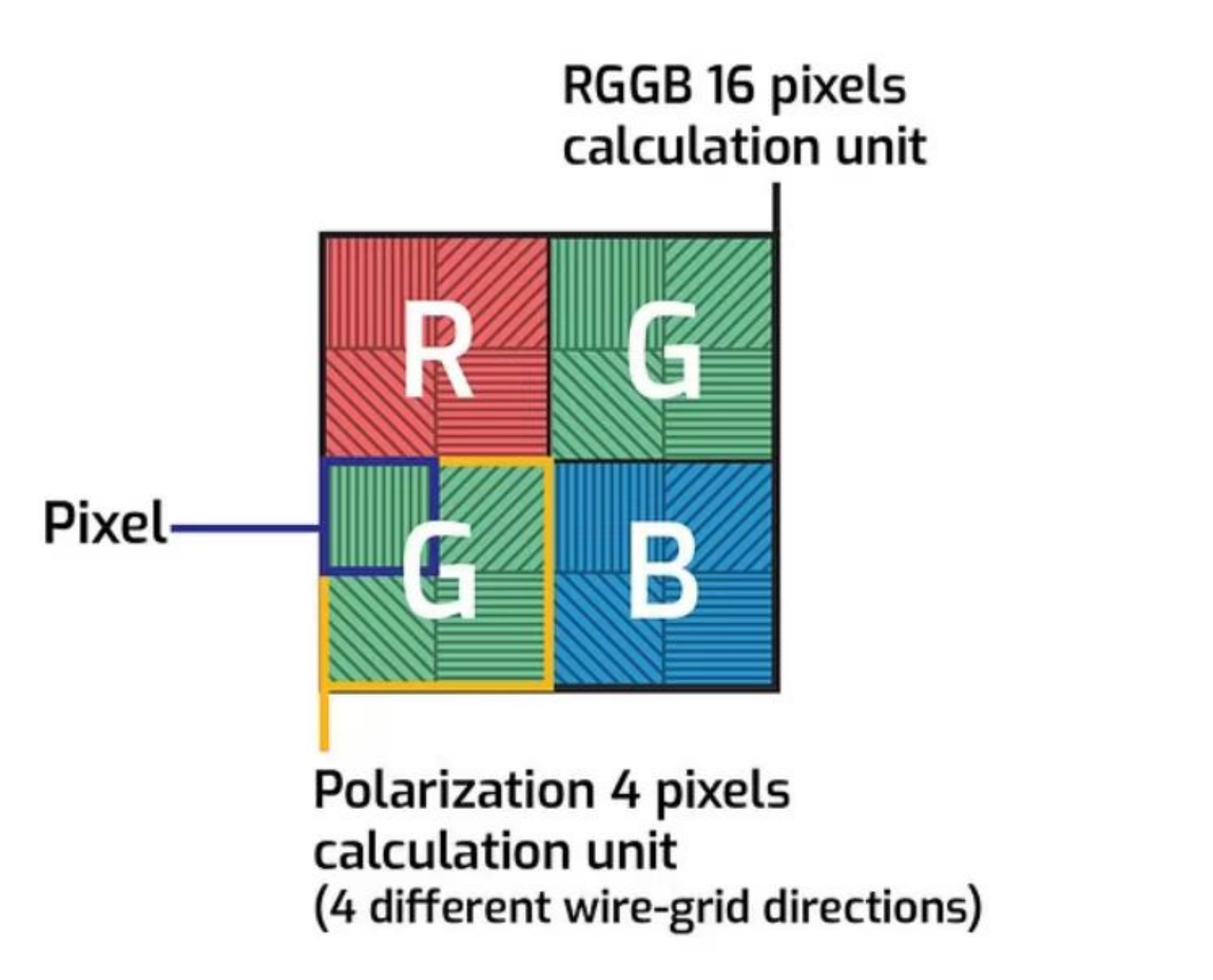}
		\end{minipage}%
	}
	\hspace{0.015\textwidth}
	\subfloat[Illumination distribution of the 10,000 RGB-P Wild image sets.\label{fig6b}]{%
		\begin{minipage}[b]{0.43\textwidth}
			\centering
			\includegraphics[width=1\linewidth]{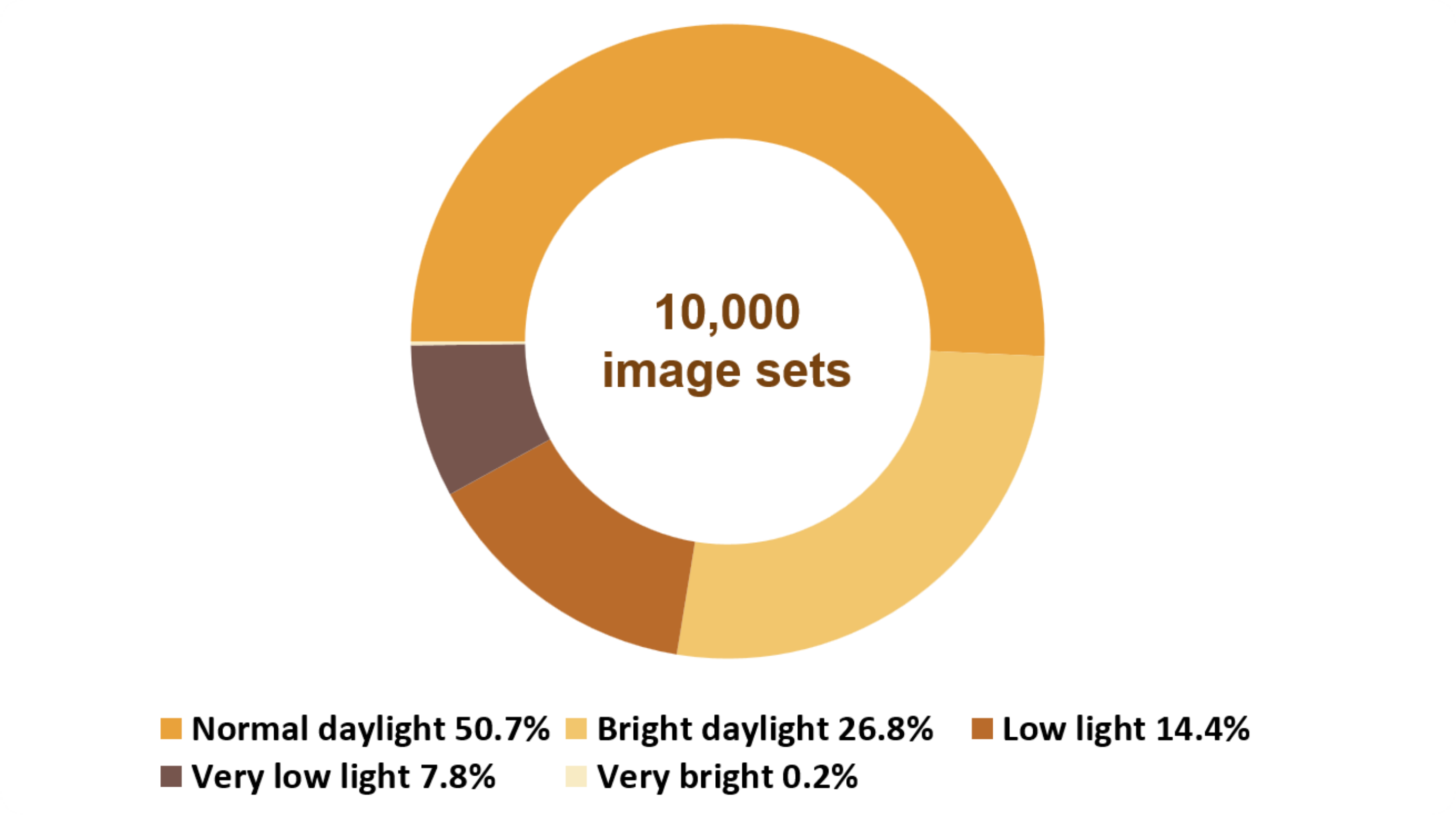}
		\end{minipage}%
	}
	\caption{Acquisition characteristics of RGB-P Wild: (a) color-polarization sampling and (b) natural-illumination distribution.}
	\label{fig6}
\end{figure*}

RGB-P Wild was collected mainly in outdoor campus environments. As shown in \cref{fig6b}, the automatic image-brightness groups contain 781 very-low-, 3,941 low-, 5,070 medium-, 184 high-, and 24 very-high-brightness image sets, corresponding to 7.81\%, 39.41\%, 50.70\%, 1.84\%, and 0.24\%, respectively. These appearance-based groups characterize the illumination diversity of the dataset and should not be interpreted as weather annotations.

\noindent\textbf{Capture Setup.} Images were acquired using a LUCID TRI050S1-QC polarization camera incorporating Sony Polarsens technology. As illustrated in \cref{fig6a}, its pixel-level color-filter array is integrated with micro-polarizers at four linear orientations ($0^\circ$, $45^\circ$, $90^\circ$, and $135^\circ$), enabling the acquisition of color-resolved linear-polarization observations \cite{Alpher84}.

The camera is operated via specialized Arenaview software for data adjustment and image acquisition. We maintain consistent exposure settings of 47,178.6 microseconds across all captures. The RGB images are captured at a resolution of $2448 \times 2048$, while each polarization image is recorded at a resolution of $1224 \times 1024$.

Our framework predicts four linear polarization orientations and does not estimate circular polarization.

\begin{table*}[h]
	\centering
	\caption{Comparison of representative polarization datasets. RGB-P Wild provides large-scale real-world paired RGB-polarization samples at the scene level, covering diverse materials and illumination conditions that are useful for RGB-to-polarization generation and vehicle-relevant reflective-scene perception studies.}
	\label{tab1}
	\begin{tabular}{l|c|c|c|c|c|c}
		\toprule
		\midrule
		Dataset & Size & Resolution & Collection  & Public & Level & Environment \\ \hline
		RGB-P Glass\cite{Alpher06} & 4511 & $612 \times 512$ & real-world & \checkmark & Scene & Indoor \\  \hline
		SPW\cite{Alpher05} & 522 & $1224 \times 1024$ & real-world & \checkmark & Scene & Outdoor \\ \hline 
		DeepSfP\cite{Alpher37} & 300 & $256 \times 256$ & real-world & \checkmark & Object & Outdoor \\ \hline
		Spectro-Polarimetric Dataset\cite{Alpher01} & 2022/311 & $1900 \times 2100$/$512 \times 612$ & real-world & \checkmark & Scene & Outdoor \\ \hline
		Kondo\cite{Alpher42} & 44305 & $256 \times 192$ & synthetic & $\times$ & Object & Outdoor \\ \hline
		PolarAnything\cite{Alpher65} & 1148 & $1224 \times 1024$ & real-world & \checkmark & Scene & Outdoor \\ \hline
		Ours & 10000 & $1224 \times 1024$ & real-world & \checkmark & Scene & Outdoor \\ \hline
		\bottomrule
	\end{tabular}
\end{table*}

\subsection{Implementation Details}

\indent We implemented PolarVis using PyTorch and trained on an NVIDIA GeForce RTX 3090 GPU. The model was trained with a batch size of 2, optimized using the Adam algorithm\cite{Alpher38} with a learning rate set to $5 \times 10^{-4}$. Training was conducted over 100 epochs, with checkpoints saved every 10 epochs. The final reported reconstruction metrics were computed on the held-out 2,500-image test split. To maintain reproducibility, a random seed of 42 was applied. The input resolution was standardized to $512 \times 512$ across the entire dataset, which was split into 7,500 images for training and 2,500 for testing. The model achieved convergence within approximately 60 hours on a single RTX 3090 GPU.

Reconstruction quality is evaluated using normalized-domain MSE \cite{Alpher39,Alpher40}, pixel-domain MSE under the original protocol, PSNR, and SSIM. Because channel normalization changes channel scales non-uniformly, normalized and pixel-domain MSE are reported separately and are not converted using a single factor of $255^2$.

\subsection{Vehicle-Oriented Cross-Dataset Evaluation}

\begin{figure*}[!t]
	\centering
	\begin{minipage}[t]{\textwidth}
		\centering
		\includegraphics[width=\linewidth]{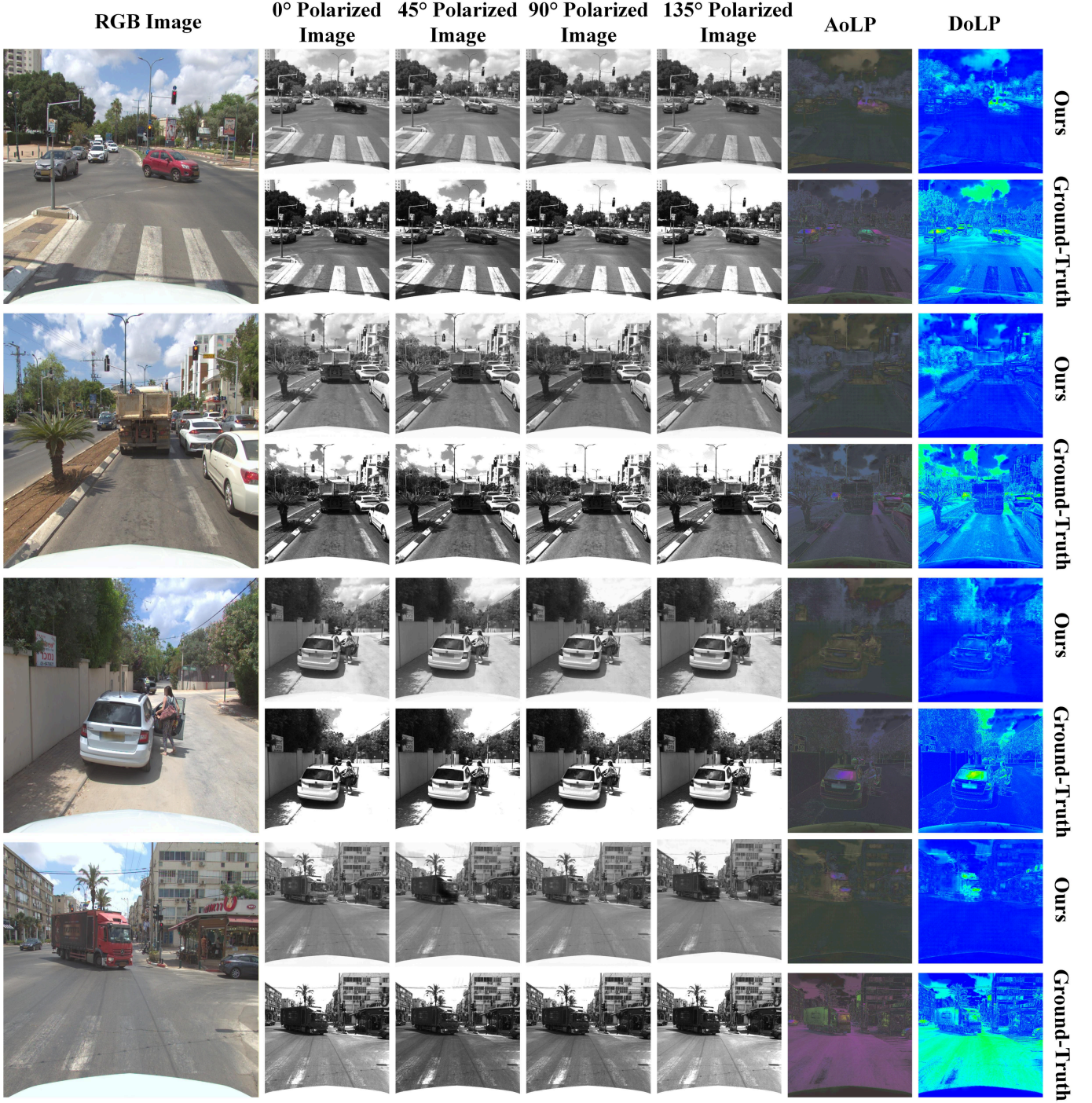}
		\caption{Qualitative transfer to the General Motors Polarimetric Imaging for Perception dataset \cite{Alpher43}. PolarVis predicts four-angle images from RGB, from which AoLP and DoLP are derived. Reference four-angle images are reconstructed using \cref{eq1.22}, not directly measured; reference AoLP and DoLP are provided by the dataset.}
		\label{fig7}
	\end{minipage}
\end{figure*}

We evaluate transfer to the General Motors Polarimetric Imaging for Perception dataset \cite{Alpher43} through qualitative visualization and image-level polarization statistics. PolarVis receives the released RGB image and predicts four-angle images; their AoLP and DoLP are computed using \cref{eq1.20,eq1.21}. No GM samples are used for training, fine-tuning, or checkpoint selection. Because raw four-angle measurements are unavailable, reference angle images are reconstructed as
\begin{equation}
	\label{eq1.22}
	\begin{aligned}
		I_{\theta}^{\mathrm{ref}} &= \frac{L_{\mathrm{RGB}}}{2}
		\Big[1+\mathrm{DoLP}^{\mathrm{ref}}
		\cos\big(2\theta-2\mathrm{AoLP}^{\mathrm{ref}}\big)\Big], \\
		\theta &\in \{0^\circ,45^\circ,90^\circ,135^\circ\}.
	\end{aligned}
\end{equation}
Here, $L_{\mathrm{RGB}}$ is an RGB-derived luminance proxy, not a measured Stokes intensity. The reconstructed four-angle references are therefore used only for visualization. In contrast, the AoLP and DoLP maps released with the GM dataset are treated as ground truth for the image-level comparison in \cref{tab:polar_frame_compare}. Representative qualitative results are shown in \cref{fig7}, while quantitative vehicle detection is evaluated separately on RGBP-Car.

As shown in \cref{fig7}, the generated angle images retain the dominant road, vehicle, lane, vegetation, and sky structures under the cross-dataset shift. Differences remain in the magnitude and spatial distribution of the predicted polarization attributes. The image-level statistics in \cref{tab:polar_frame_compare} show that the predicted DoLP mean is lower than the GM ground-truth mean for all four selected frames. These statistics summarize the global attribute distributions and are interpreted together with the qualitative maps; they do not replace pixel-wise error metrics.

\begin{table*}[t]
	\centering
	\caption{Image-level comparison of GM ground-truth and PolarVis-predicted polarization attributes. AoLP is summarized by the axial circular mean over the common mask defined by ground-truth DoLP greater than 0.05, and DoLP is averaged over all pixels. These descriptive statistics do not replace pixel-wise error metrics.}
	\label{tab:polar_frame_compare}
	\begin{tabular}{c|c|c|c}
		\toprule
		\midrule
		GT AoLP ($^\circ$) & PolarVis AoLP ($^\circ$) & GT DoLP & PolarVis DoLP \\ \hline
		117.5878 & 106.1014 & 0.1361 & 0.0929 \\ \hline
		110.9761 & 101.2570 & 0.1387 & 0.1026 \\ \hline
		114.4993 & 106.8212 & 0.0836 & 0.0773 \\ \hline
		148.3800 & 110.9311 & 0.1824 & 0.1146 \\ \hline
		\bottomrule
	\end{tabular}
\end{table*}

\begin{figure*}[!t]
	\centering
	\begin{minipage}[t]{\textwidth}
		\centering
		\includegraphics[width=\linewidth,height=0.45\textheight,keepaspectratio]{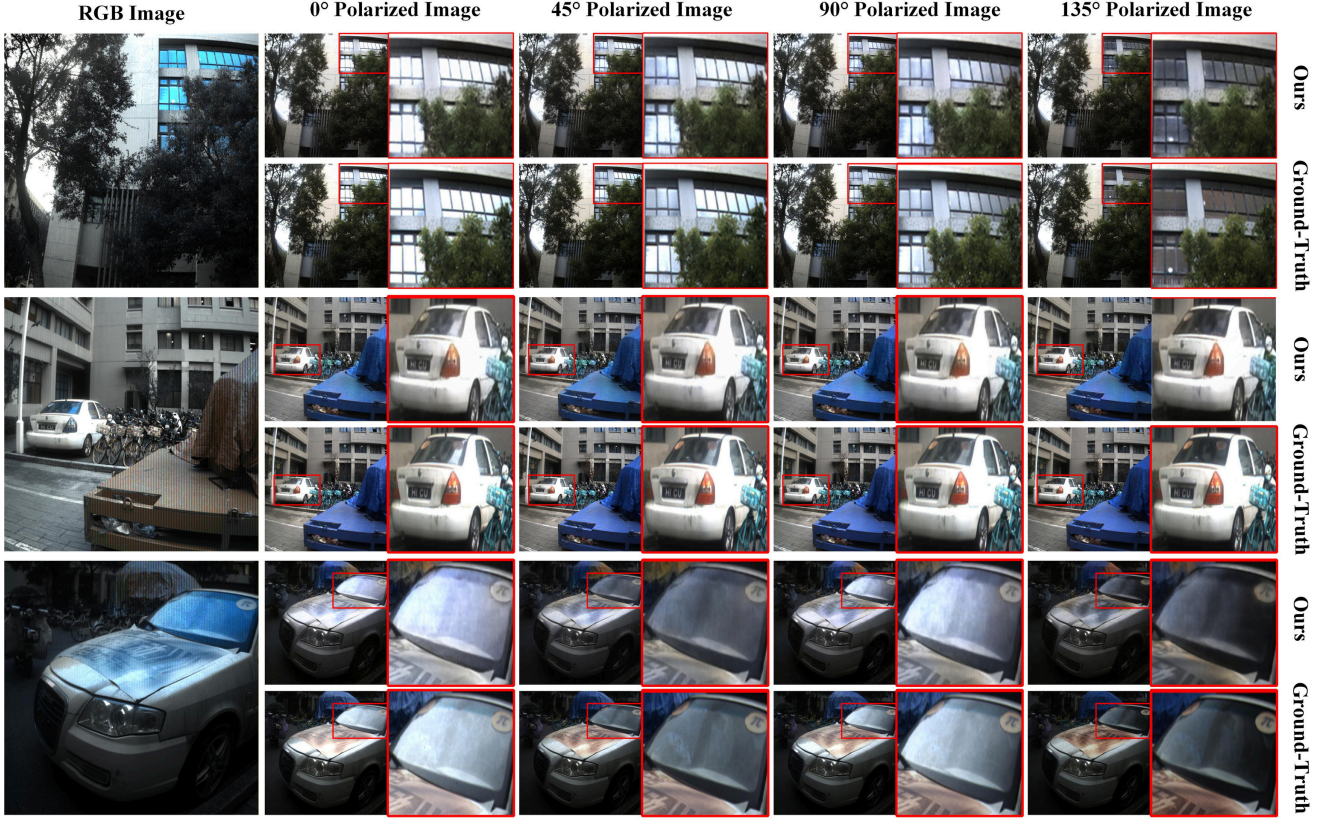}
		\caption{Supplementary in-domain qualitative comparison on RGB-P Wild. Complementing the vehicle-scene cross-dataset evaluation in \cref{fig7}, PolarVis-generated polarization images are compared with direct ground-truth polarization observations from RGB-P Wild. The generated results preserve the main structural appearance and polarization-related intensity patterns across diverse reflective materials and illumination conditions, while some fine local differences remain in highly reflective or low-texture regions.}
		\label{fig8}
	\end{minipage}
\end{figure*}

\subsection{RGB-to-Polarization Generation Performance on RGB-P Wild}

For complementary in-domain analysis, \cref{fig8} compares PolarVis outputs with direct four-angle polarization observations from the RGB-P Wild test split. Together with the vehicle-mounted cross-dataset comparison in \cref{fig7}, these results distinguish transfer to a driving-oriented domain from reconstruction performance within the diverse reflective-scene distribution represented by RGB-P Wild.

\noindent\textbf{Diffusion Model Analysis.} We compared our simplified diffusion model to the conventional approach\cite{Alpher23}, incorporating several key modifications. Notably, we integrated direct time embedding and simplified the computation of posterior variance using \cref{eq1.9}.

Additionally, we reduced the number of diffusion steps from 1,000 to 100 and streamlined the noise processing. Experimental results indicate that our optimized diffusion module improves the RGB-to-polarization reconstruction quality across multiple metrics: normalized MSE is reduced from 1.5879 to 0.0465 (a decrease of approximately 1.54), pixel-domain MSE is reduced from 5273.0545 to 155.0706 (a reduction of over 5100), PSNR is improved from 11.51 to 27.83 (an increase of around 16.32), and SSIM is enhanced from 0.3686 to 0.8597 (a gain of approximately 0.49), as shown in \cref{tab2}. In contrast, the conventional diffusion model displays prohibitively high MSE values and low PSNR/SSIM scores, struggling to generate meaningful polarization images. This suggests that the excessive noise introduced by the conventional setting can corrupt polarization-related structural information, thereby affecting convergence and detail recovery.

\noindent\textbf{Decoder Comparison.} We conducted a comparative analysis of our enhanced decoder against two prominent models: Rombach's decoder\cite{Alpher16}, optimized for multi-task processing, and the distinctive decoder architecture of StarGAN v2\cite{Alpher41}.

When compared to Rombach et al.'s decoder, our normalized MSE is reduced from 0.0499 to 0.0465 (a decrease of approximately 0.0034), pixel-domain MSE is reduced from 165.4929 to 155.0706 (a reduction of around 10.42), PSNR is improved from 27.65 to 27.83 (an increase of 0.18), and SSIM is enhanced from 0.8434 to 0.8597 (a gain of approximately 0.0163). This suggests that the techniques employed in our decoder design, such as attention mechanisms and residual connections, are more effective in capturing and integrating multi-scale features, thereby enhancing reconstruction accuracy. In comparison to the decoder used in StarGAN v2, our design achieves a normalized MSE reduction from 0.0608 to 0.0465 (a decrease of approximately 0.0143), a pixel-domain MSE reduction from 202.0217 to 155.0706 (a reduction of around 46.95), a PSNR increase from 26.28 to 27.83 (an improvement of 1.55), and an SSIM gain from 0.8018 to 0.8597 (an enhancement of approximately 0.0579). These results indicate that our decoder is more suitable for polarization image reconstruction tasks and exhibits improved generalization performance. Overall, the proposed decoder improves all objective metrics (MSE, PSNR, and SSIM) in this setting, which can be attributed to the combination of multi-scale feature fusion and attention-based feature recalibration.

\begin{table*}
	\centering
	\caption{Quantitative comparison of different diffusion and decoder configurations on RGB-P Wild. The proposed configuration achieves the lowest MSE and the highest PSNR/SSIM among the compared settings.}
	\label{tab2}
	\begin{tabular}{l|>{\arraybackslash}m{3.1cm}|>{\arraybackslash}m{3.3cm}|>{\arraybackslash}m{2.4cm}|>{\arraybackslash}m{2.3cm}}
		\toprule
		\midrule
		Method &
		Mean Squared Error (MSE) (normalized) $\downarrow$ &
		Mean Squared Error (MSE) $\downarrow$ &
		Peak Signal-to-Noise Ratio (PSNR) $\uparrow$ &
		Structural Similarity Index (SSIM) $\uparrow$ \\ \hline
		Traditional diffusion model & 1.5879 & 5273.0545 & 11.51 & 0.3686 \\ \hline
		Rombach's decoder\cite{Alpher16} & 0.0499 & 165.4929 & 27.65 & 0.8434 \\ \hline
		StarGAN v2's decoder\cite{Alpher41} & 0.0608 & 202.0217 & 26.28 & 0.8018 \\ \hline
		Ours & 0.0465 & 155.0706 & 27.83 & 0.8597 \\ \hline
		\bottomrule
	\end{tabular}
\end{table*}

\begin{figure*}[!t]
	\centering
	\begin{minipage}[t]{\textwidth}
		\centering
		\includegraphics[width=\linewidth,height=0.45\textheight,keepaspectratio]{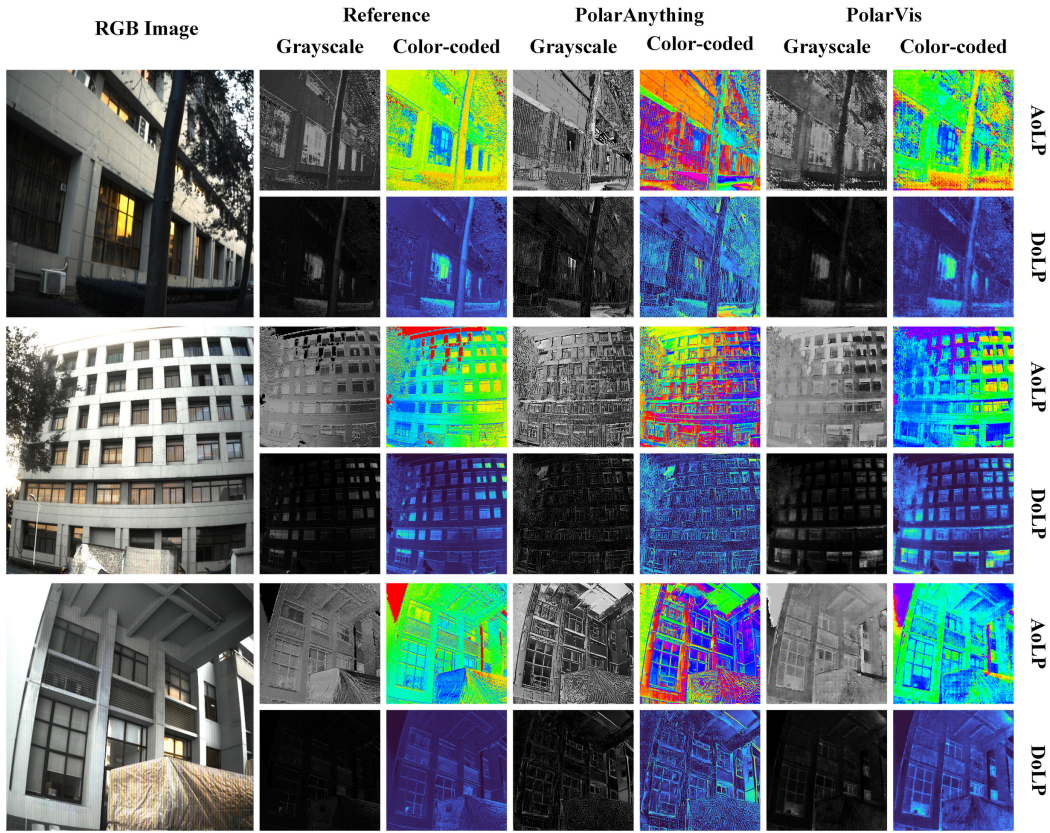}
		\caption{Qualitative comparison with PolarAnything \cite{Alpher65} on RGB-P Wild. Each scene shows the RGB input and reference, PolarAnything, and PolarVis AoLP/DoLP maps. Reference maps are derived from measured four-angle images. Grayscale and color-coded views are shown for each source; each pair displays the same underlying attribute. Quantitative results are reported in \cref{tab5}.}
		\label{fig12}
	\end{minipage}
\end{figure*}

\noindent\textbf{Comparison with PolarAnything.} We evaluate the released pretrained PolarAnything model \cite{Alpher65} and PolarVis on the 2,500-image RGB-P Wild evaluation split. Reference AoLP and DoLP are computed from the measured four-angle images. The predictions are compared visually in \cref{fig12} and quantitatively in \cref{tab5}. PolarVis obtains an AoLP error of $25.66^\circ$ and DoLP MAE/RMSE of 0.0404/0.0614, compared with $46.03^\circ$ and 0.1084/0.1528 for PolarAnything. PolarAnything uses its released weights without RGB-P Wild fine-tuning, whereas PolarVis is trained on RGB-P Wild; the comparison therefore does not assume matched training conditions.

The controlled reconstruction in \cref{tab5}(b) supplies the same reference $S_0$ to both methods. PolarVis achieves MAE/PSNR/SSIM of 0.01529/33.11~dB/0.9737, compared with 0.03469/25.19~dB/0.8385 for PolarAnything. This auxiliary analysis isolates predicted polarimetric attributes and is not an end-to-end inference result.

AoLP errors use the circular distance $\min(\Delta,\pi-\Delta)$, where $\Delta=|\operatorname{mod}(a,\pi)-\operatorname{mod}(a^{\mathrm{ref}},\pi)|$. Evaluation uses the common valid-pixel mask with reference DoLP greater than 0.05. DoLP MAE and RMSE are computed in the dimensionless attribute domain. The controlled reconstruction metrics use grayscale images, and the reported metrics are averaged over images.

\begin{table}[t]
	\centering
	\caption{Polarimetric comparison on the 2,500-image RGB-P Wild evaluation split: (a) AoLP/DoLP errors; (b) four-angle reconstruction with a common reference $S_0$. Part (b) is a controlled analysis, not end-to-end inference.}
	\label{tab5}
	\setlength{\tabcolsep}{5.4pt}
	\begin{tabular}{l|c|c|c}
		\toprule
		\midrule
		\multicolumn{4}{c}{(a) AoLP/DoLP evaluation} \\ \hline
		Method & AoLP error ($^\circ$) $\downarrow$ & DoLP MAE $\downarrow$ & DoLP RMSE $\downarrow$ \\ \hline
		PolarAnything & 46.03 & 0.1084 & 0.1528 \\ \hline
		PolarVis & 25.66 & 0.0404 & 0.0614 \\ \hline
		\multicolumn{4}{c}{(b) Controlled reconstruction with reference $S_0$} \\ \hline
		Method & Four-angle MAE $\downarrow$ & PSNR (dB) $\uparrow$ & SSIM $\uparrow$ \\ \hline
		PolarAnything & 0.03469 & 25.19 & 0.8385 \\ \hline
		PolarVis & 0.01529 & 33.11 & 0.9737 \\ \hline
		\bottomrule
	\end{tabular}
\end{table}		

\noindent\textbf{Efficiency.} As reported in \cref{tab7}, PolarVis contains 157.35 million parameters and requires 103.82~ms per $512\times512$ image with 10 refinement steps on an RTX~4090, corresponding to 9.63~FPS. The measured latency excludes file I/O, polarization conversion, and downstream inference. The table reports speed at different step counts; it does not establish their relative reconstruction quality.

\begin{table}[t]
	\centering
	\caption{PolarVis inference efficiency at $512\times512$ resolution on an NVIDIA RTX~4090. The model contains 157.35 million parameters. Latency excludes file I/O, polarization conversion, and downstream inference.}
	\label{tab7}
	\begin{tabular}{l|c|c|c}
		\toprule
		\midrule
		Steps & Latency (ms) $\downarrow$ & FPS $\uparrow$ & Peak memory (MB) $\downarrow$ \\ \hline
		1 & 88.74 & 11.27 & 2682.76 \\ \hline
		5 & 95.19 & 10.51 & 2683.17 \\ \hline
		10 & 103.82 & 9.63 & 2682.76 \\ \hline
		\bottomrule
	\end{tabular}
\end{table}

\subsection{Ablation Studies}

\noindent\textbf{Channel Attention Mechanism Analysis.}  We performed ablation experiments to isolate the effect of channel attention, comparing models with channel-only attention against our complete decoder integrated with the FFA module. Key modifications involved retaining channel-wise feature screening while introducing the FFA module’s multi-scale fusion and attention recalibration. Results indicate: normalized MSE decreases from 0.0567 to 0.0465 (reduction of 0.0102), pixel-domain MSE drops from 187.8577 to 155.0706 (reduction of 32.7871), PSNR improves from 26.79 to 27.83 (gain of 1.04), and SSIM enhances from 0.8131 to 0.8597 (gain of 0.0466), as detailed in \cref{tab3}.  

The channel-only attention mechanism, though effective for initial feature filtering, lacks multi-scale interaction, resulting in suboptimal feature utilization. In contrast, the FFA module complements channel attention by integrating cross-scale information, reducing redundancy, and enhancing polarization feature fidelity—critical for accurate image reconstruction.  

\noindent\textbf{Spatial Attention Mechanism Analysis.}  We assessed the role of spatial attention by contrasting models with spatial-only attention against our FFA-integrated decoder.  Our design retains spatial correlation capture but augments it with the FFA module’s multi-dimensional feature alignment. Experimental outcomes show: normalized MSE reduces from 0.1769 to 0.0465 (reduction of 0.1304), pixel-domain MSE falls from 588.6359 to 155.0706 (reduction of 433.5653), PSNR improves from 21.30 to 27.83 (gain of 6.53), and SSIM enhances from 0.5939 to 0.8597 (gain of 0.2658), referenced in \cref{tab3}.  

Spatial-only attention struggles to capture global feature relationships due to its local focus, leading to fragmented reconstructions. The FFA module addresses this by enforcing multi-scale spatial coherence, enabling robust polarization feature integration and superior detail recovery—essential for high-quality image generation.  

\noindent\textbf{Single-scale Convolution Analysis.}  We analyzed the impact of convolution scale by comparing models with $3 \times 3$ convolution only (removing multi-scale $1 \times 1$ and $5 \times 5$ convolutions) against our FFA-equipped decoder.  By restricting to a single convolution scale, we observed: normalized MSE increases from 0.0465 to 1.6987 (increase of 1.6522), pixel-domain MSE rises from 155.0706 to 5644.3328 (increase of 5489.2622), PSNR drops from 27.83 to 11.38 (loss of 16.45), and SSIM falls from 0.8597 to 0.3743 (loss of 0.4854), as shown in \cref{tab3}.  

A single $3 \times 3$ convolution fails to adapt to the complex distribution of polarization features, causing severe information loss. The FFA module’s multi-scale convolutions ($1 \times 1$, $3 \times 3$, $5 \times 5$) provide complementary receptive fields, capturing both fine details and global context-enabling comprehensive polarization feature learning and reconstruction.  

Overall, the ablation results indicate that the FFA module contributes to polarization image reconstruction by combining multi-scale convolutional branches with channel-spatial feature recalibration. Compared with single-mechanism variants, the full module achieves lower reconstruction errors and better structural similarity, suggesting that both cross-scale feature aggregation and attention-based recalibration are useful for this task.

\begin{table*}
	\centering
	\caption{Ablation study of the Multi-Scale Feature Fusion and Attention Feature Recalibration (FFA) module on RGB-P Wild. The comparison is conducted in terms of normalized Mean Squared Error (MSE), Mean Squared Error (MSE), Peak Signal-to-Noise Ratio (PSNR), and Structural Similarity Index (SSIM). The results show the contribution of multi-scale feature fusion and attention-based recalibration to polarization image reconstruction.}
	\label{tab3}
	\begin{tabular}{l|>{\arraybackslash}m{3.1cm}|>{\arraybackslash}m{3.3cm}|>{\arraybackslash}m{2.4cm}|>{\arraybackslash}m{2.3cm}}
		\toprule
		\midrule
		Method &
		Mean Squared Error (MSE) (normalized) $\downarrow$ &
		Mean Squared Error (MSE) $\downarrow$ &
		Peak Signal-to-Noise Ratio (PSNR) $\uparrow$ &
		Structural Similarity Index (SSIM) $\uparrow$ \\ \hline
		Only channel attention mechanism & 0.0567 & 187.8577 & 26.79 & 0.8131 \\ \hline
		Only spatial attention mechanism & 0.1769 & 588.6359 & 21.30 & 0.5939 \\ \hline
		\makecell[l]{Only $3 \times 3$ convolution layer \\ (remove $1 \times 1$ and $5 \times 5$ convolutions)}
		& 1.6987 & 5644.3328 & 11.38 & 0.3743 \\ \hline
		Full & 0.0465 & 155.0706 & 27.83 & 0.8597 \\ \hline
		\bottomrule
	\end{tabular}
\end{table*}

\section{Downstream Evaluation}

\subsection{Transparent and Reflective Object Segmentation}

We first use PGSNet \cite{Alpher06} and RGB-P Glass \cite{Alpher06,Alpher09} as a proxy evaluation for reflective and transparent regions. RGB images are processed by PolarVis, and AoLP/DoLP derived from the predicted angle images are supplied to the existing segmentation pipeline. RGB-P Glass is not a driving benchmark; this experiment evaluates whether generated polarization retains information useful for glass-like boundary perception.

We compute
\begin{align}
	S_0 &= \tfrac{1}{2}(I_0+I_{45}+I_{90}+I_{135}), \\
	S_1 &= I_0-I_{90}, \qquad S_2=I_{45}-I_{135}, \\
	\operatorname{AoLP} &= \tfrac{1}{2}\operatorname{atan2}(S_2,S_1)\bmod\pi, \label{eq1.20} \\
	\operatorname{DoLP} &= \frac{\sqrt{S_1^2+S_2^2}}{\max(S_0,\epsilon)}. \label{eq1.21}
\end{align}
The two-argument $\operatorname{atan2}$ preserves the angular quadrant, while modulo $\pi$ accounts for the periodicity of linear polarization.

\begin{figure*}[ht]
	\centering
	\begin{minipage}[t]{\textwidth}
		\centering
		\includegraphics[width=\linewidth]{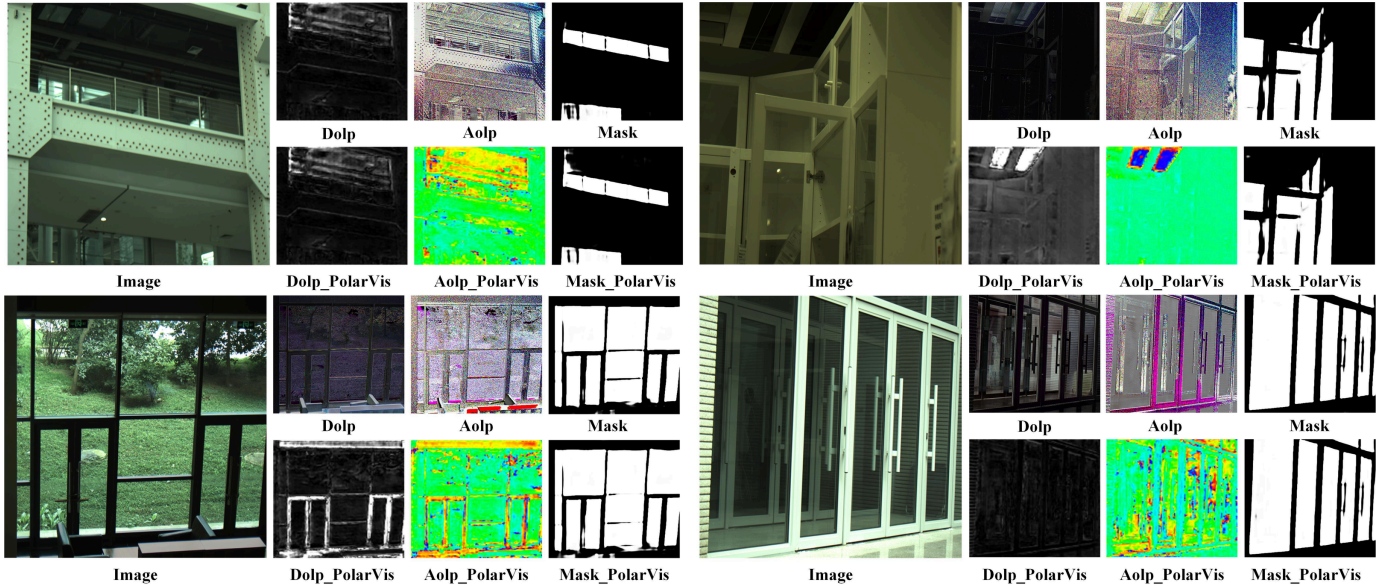}
		\caption{Transparent and reflective object segmentation results using real and PolarVis-generated polarization cues. The figure visualizes RGB images, real DoLP/AoLP maps, PolarVis-generated DoLP/AoLP maps, and segmentation masks. The results indicate that generated polarization cues can provide useful boundary and reflection-related information for glass-like scene perception.}
		\label{fig9}
	\end{minipage}
\end{figure*}

Predicted masks are evaluated against ground-truth labels using intersection over union (IoU), weighted F-measure ($F_\beta^w$), mean absolute error (MAE), and balanced error rate (BER). Qualitative results and quantitative comparisons are presented in \cref{fig9,tab4}.

\begin{table*}[ht]
	\centering
	\caption{Comparison of polarization-assisted segmentation performance via PGSNet. Two input configurations are compared: (1) original RGB-P Glass inputs with real polarization information; and (2) RGB inputs paired with PolarVis-synthesized polarization information. The experiment evaluates whether generated polarization cues can support transparent and reflective object segmentation. Note: $\uparrow$ indicates higher values are better; $\downarrow$ indicates lower values are better.}
	\label{tab4}
	\begin{tabular}{l|>{\arraybackslash}m{3cm}|>{\arraybackslash}m{3cm}|>{\arraybackslash}m{2.5cm}|>{\arraybackslash}m{2.5cm}}
		\toprule
		\midrule
		PGSNet & IoU $\uparrow$ & $F_\beta^w$ $\uparrow$ & MAE $\downarrow$ & BER $\downarrow$ \\ \hline
		RGB-P Glass & 81.08 & 0.842 & 0.091 & 9.63 \\ \hline
		Generated by PolarVis & 74.11 & 0.776 & 0.140 & 13.79 \\ \hline
		\bottomrule
	\end{tabular}
\end{table*}

As shown in \cref{fig9,tab4}, generated polarization retains useful segmentation information but remains below measured polarization: IoU decreases from 81.08\% to 74.11\%, while $F_\beta^w$/MAE/BER change from 0.842/0.091/9.63\% to 0.776/0.140/13.79\%. This experiment supports transfer to transparent-object segmentation; direct vehicle detection is evaluated in the following subsection.

\subsection{Car Detection with Frozen PCDNet}

\begin{figure*}[!t]
	\centering
	\includegraphics[width=\textwidth]{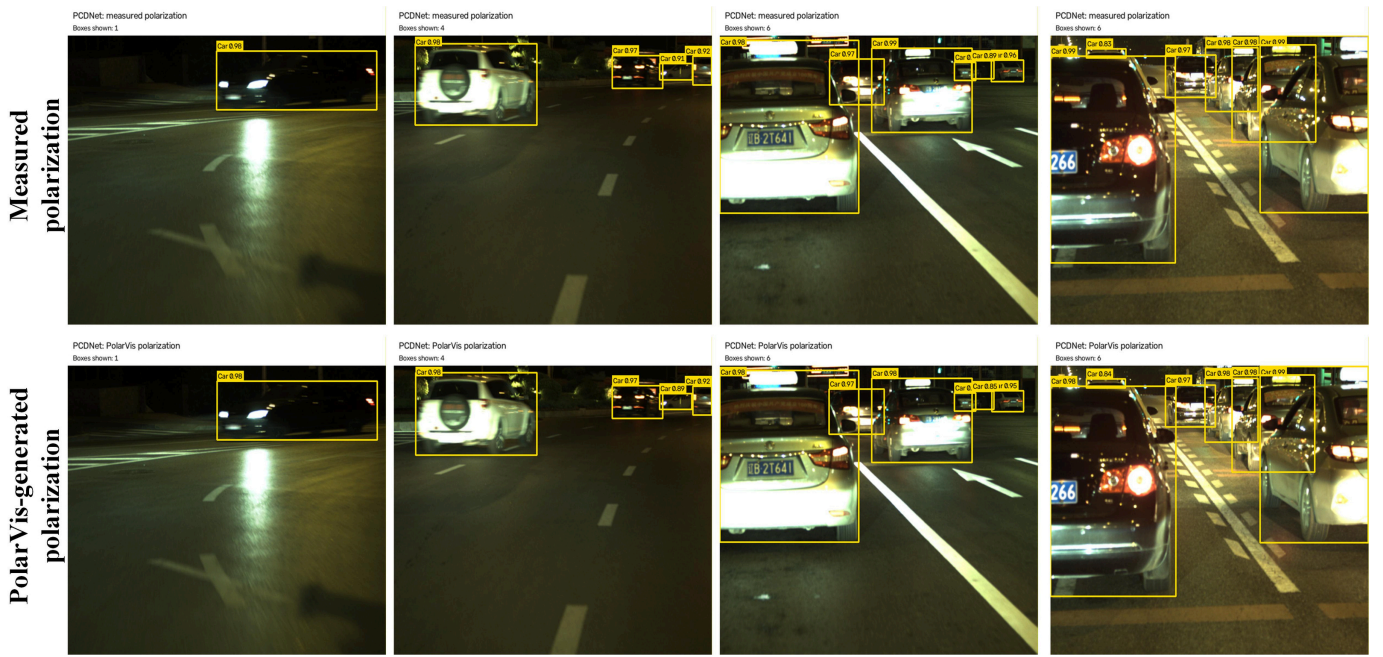}
	\caption{Vehicle detections produced by frozen PCDNet \cite{Alpher83} on four RGBP-Car examples. The top and bottom rows use measured and PolarVis-generated AoLP/DoLP, respectively. Each column uses the same RGB image and detector weights. Aggregate results are reported in \cref{tab8}.}
	\label{fig13}
\end{figure*}

We conduct a direct vehicle-oriented evaluation using RGBP-Car and the released PCDNet checkpoint \cite{Alpher83}. PolarVis is trained only on RGB-P Wild and is not fine-tuned on RGBP-Car. Two configurations are evaluated on the same 990 test images with 12,420 vehicle annotations: RGB paired with measured AoLP/DoLP, and the same RGB paired with PolarVis-generated AoLP/DoLP. Detector weights, annotations, and evaluation settings are fixed, and no detector retraining is performed. Representative paired detection results are shown in \cref{fig13}.

Predicted angle images are resized to the RGB dimensions and converted to AoLP and DoLP. AoLP is mapped to $[0,\pi)$, DoLP is clipped to $[0,1]$, and both are encoded as 8-bit three-channel TIFF inputs. This mapping is fixed before full-test evaluation. Because the exact quantization used for the released detector inputs is undocumented, the results are interpreted under this specified adapter rather than as an exact reproduction of the original data pipeline.

\begin{table}[t]
	\centering
	\caption{Vehicle detection using frozen PCDNet on 990 RGBP-Car test images. Both rows use identical detector weights and evaluation settings. Measured-polarization results are reproduced locally. All metrics are percentages.}
	\label{tab8}
	\setlength{\tabcolsep}{4.5pt}
	\begin{tabular}{l|c|c|c|c}
		\toprule
		\midrule
		Polarization input &
		P (\%) $\uparrow$ &
		R (\%) $\uparrow$ &
		AP50 (\%) $\uparrow$ &
		AP50:95 (\%) $\uparrow$ \\ \hline
		Measured & 87.6 & 82.3 & 87.0 & 59.8 \\ \hline
		PolarVis-generated & 87.6 & 81.3 & 86.3 & 59.1 \\ \hline
		\bottomrule
	\end{tabular}
\end{table}

PCDNet is evaluated at image size 640 with batch size 1, FP32 inference, a confidence threshold of 0.001, and a non-maximum suppression IoU threshold of 0.65. As shown in \cref{tab8}, measured and generated polarization yield AP50/AP50:95 values of 87.0\%/59.8\% and 86.3\%/59.1\%, respectively, corresponding to decreases of approximately 0.7 percentage points for both metrics. The qualitative results in \cref{fig13} further show that the two polarization inputs produce similar principal vehicle detections on the selected RGB scenes, although differences remain in individual boxes and confidence scores. This experiment demonstrates useful transfer to a frozen vehicle detector, but does not establish superiority over an RGB-only detector or statistical equivalence to measured polarization.

\FloatBarrier

\section{Conclusion}

This paper presented RGB-P Wild, a dataset of 10,000 paired RGB and four-angle polarization image sets, and PolarVis, a diffusion-based framework for predicting four linear-polarization images from a single RGB input. On RGB-P Wild, PolarVis achieved a normalized MSE of 0.0465, a PSNR of 27.83~dB, and an SSIM of 0.8597. The generated polarization attributes supported transparent and reflective object segmentation and retained most of the frozen PCDNet vehicle-detection performance obtained with measured polarization. Cross-dataset evaluation on GM preserved major vehicle-scene structures but also revealed differences in polarization magnitude and spatial distribution. These findings indicate that PolarVis can provide useful polarization-related cues when dedicated sensing is unavailable, while its outputs should not be treated as calibrated measurements. Future work will incorporate explicit physical-consistency constraints, collect vehicle-mounted RGB-polarization data, improve inference efficiency, and evaluate RGB-only and polarization-assisted baselines on additional driving tasks, including obstacle detection, road-surface understanding, and adverse-weather perception \cite{Alpher79,Alpher80,Alpher81,Alpher82}.

\input{main.bbl}

\begin{IEEEbiography}[{\includegraphics[width=1in,height=1.25in,clip,keepaspectratio]{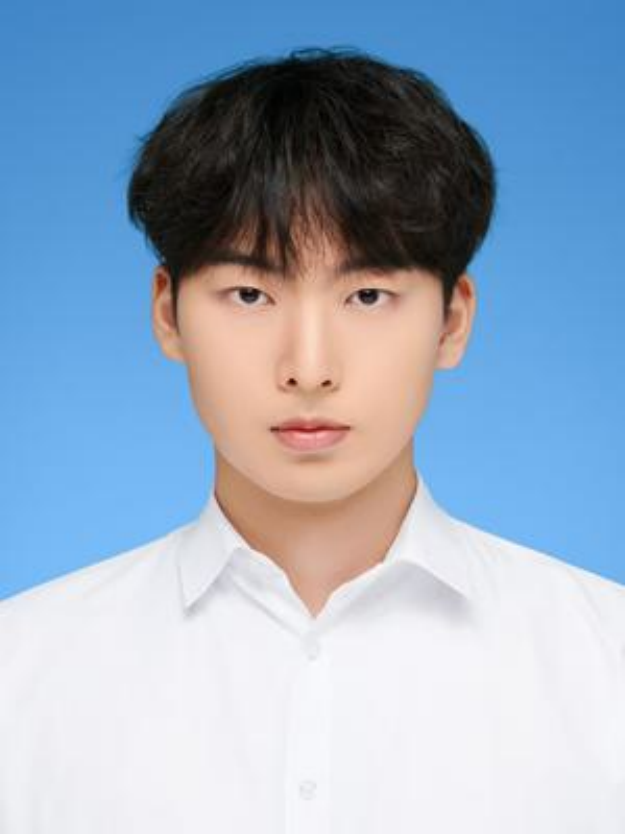}}]{Ruilin Huang}
received the B.S. degree in Communication Engineering from Jiangsu University in Zhenjiang, China, in 2023. He is currently pursuing the M.S. degree with the School of Information and Communication Engineering, Shanghai University. His research interests include professional image and video processing and computer vision.
\end{IEEEbiography}

\begin{IEEEbiography}[{\includegraphics[width=1in,height=1.25in,clip,keepaspectratio]{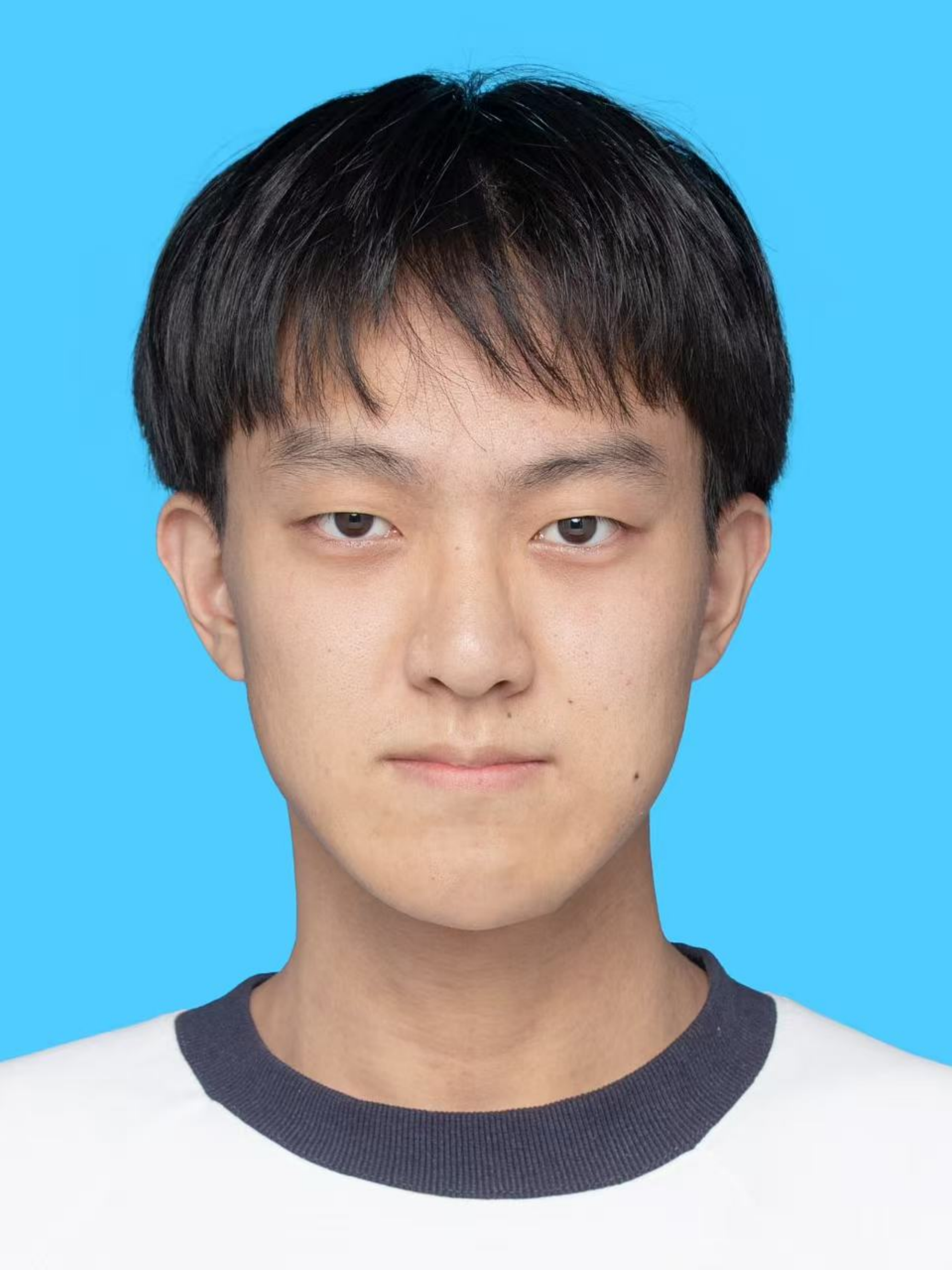}}]{Jue Wang}
received the B.S. degree in Communication Engineering from Jiangsu University in Zhenjiang, China, in 2024. He is currently pursuing the M.S. degree with the School of Information and Communication Engineering, Shanghai University. His research interests include polarization imaging in computer vision and robotic visuotactile sensing.
\end{IEEEbiography}

\begin{IEEEbiography}[{\includegraphics[width=1in,height=1.25in,clip,keepaspectratio]{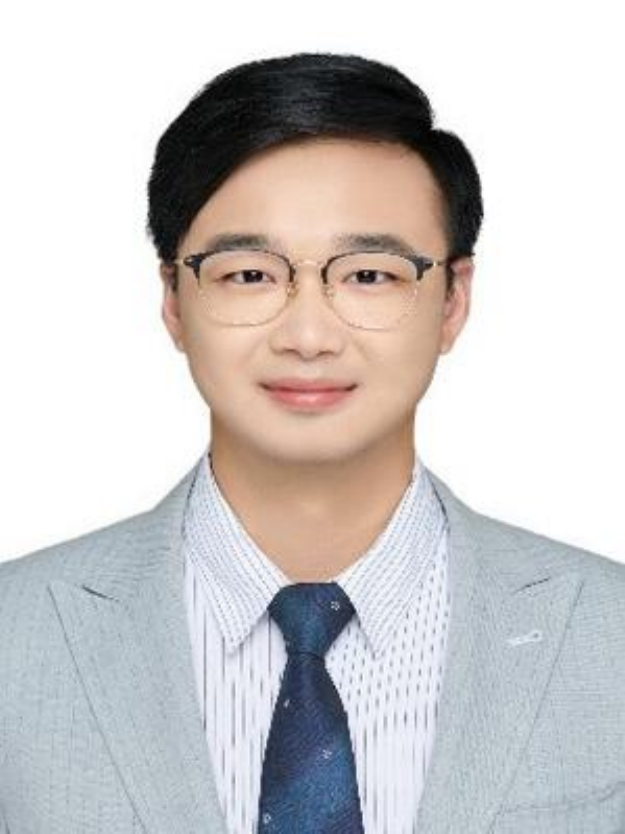}}]{Zhen Tian}
received the B.E. degree in electrical engineering and automation in 2013, the M.E. degree in control science and engineering in 2016, and the Ph.D. degree in computer science and technology in 2020, all from Chongqing University of Posts and Telecommunications, Chongqing, China. From September 2018 to September 2019, he was a visiting Ph.D. student at McMaster University, Hamilton, Canada. He was a postdoctoral researcher at Huawei Technologies Co., Ltd. from June 2021 to June 2023. Since July 2023, he has been an Associate Researcher at the Guangdong Laboratory of Artificial Intelligence and Digital Economy (SZ). His current research interests include information fusion, embodied AI, and autonomous driving.
\end{IEEEbiography}

\begin{IEEEbiography}[{\includegraphics[width=1in,height=1.25in,clip,keepaspectratio]{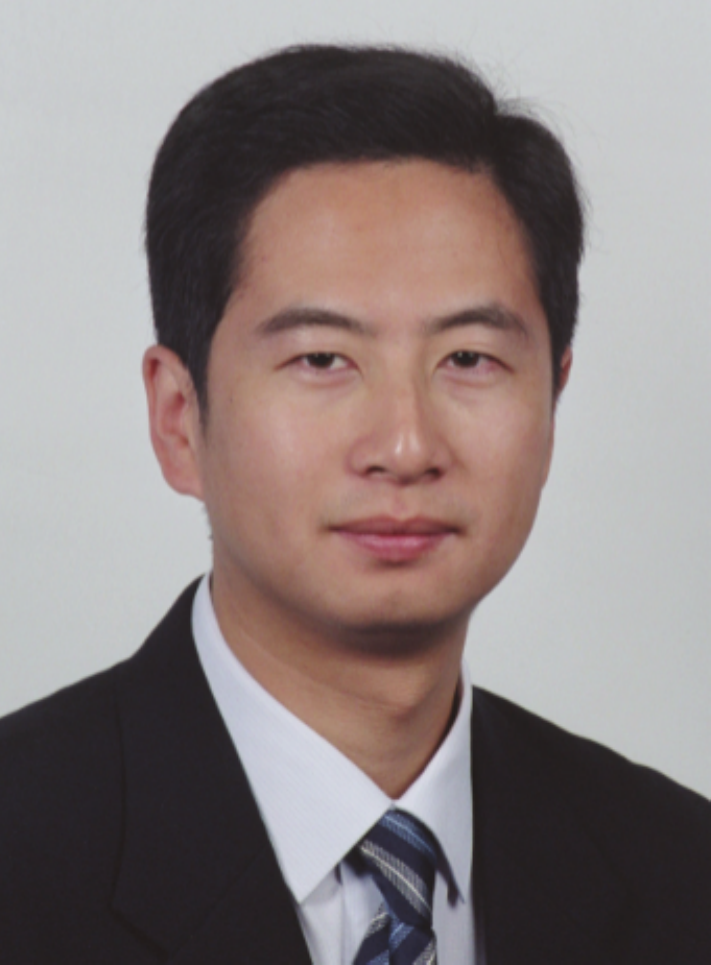}}]{F. Richard Yu}
(Fellow, IEEE) obtained his PhD in Electrical Engineering from the University of British Columbia (UBC) in 2003. His research interests include intelligent and autonomous systems, the Internet of Things (IoT), machine learning, and blockchain. He has been consistently recognized as a Clarivate “Highly Cited Researcher,” Stanford “Top 2\% Most Cited Scientist,” and ScholarGPS “Top 0.05\% Highly Ranked Scholar” for several consecutive years. He has received several Best Paper Awards from first-tier conferences, two Carleton Research Achievement Awards (2012 and 2021), and the Ontario Early Researcher Award in 2011. He is a Member of the Academia Europaea (MAE) and a Fellow of the Royal Society of Canada (FRSC), Canadian Academy of Engineering (CAE), the Engineering Institute of Canada (EIC), IEEE, and IET. He also serves as a Board Member of the IEEE Vehicular Technology Society and as an IEEE Distinguished Lecturer.
\end{IEEEbiography}

\begin{IEEEbiography}[{\includegraphics[width=1in,height=1.25in,clip,keepaspectratio]{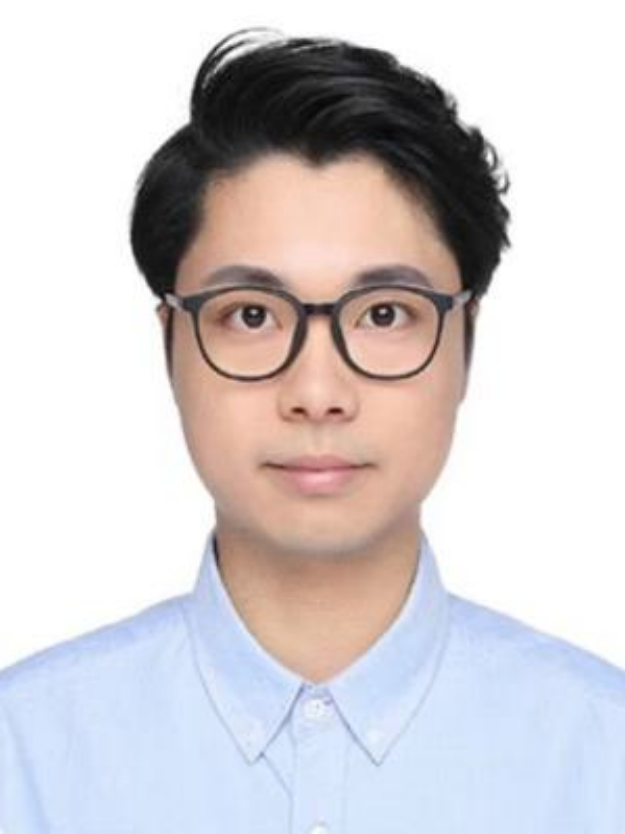}}]{Xia Hua}
received the Ph.D. degree from the School of Electronic Science and Engineering, Nanjing University, Nanjing, China. He was a Post-doctoral Researcher at Huawei Technologies Co., Ltd. Currently, he is a Lecturer at Shanghai University, with a specialization in computational vision. His research focuses on intelligent visual information processing, integrating optics, signal processing, and machine learning to achieve miniature and intelligent multi-dimensional and multi-scale visual information capture and reconstruction. He participated in the General Program of the National Natural Science Foundation of China titled “Research on Fast Imaging Methods in Lens-free Microscopy” and has published papers in journals like Nature Communications and Advanced Photonics.
\end{IEEEbiography}

\vfill

\end{document}